\documentclass[12pt]{article}

\usepackage[a4paper,margin=1in]{geometry}
\usepackage{amsmath}              
\usepackage{newtxtext}
\usepackage{newtxmath}
\usepackage{setspace}
\usepackage{microtype}

\usepackage{graphicx}
\usepackage{booktabs}
\usepackage{tabularx}
\usepackage{array}
\usepackage{makecell}
\usepackage{multirow}
\usepackage{caption}
\usepackage{float}
\usepackage{placeins}

\newcolumntype{Y}{>{\raggedright\arraybackslash}X}
\newcolumntype{L}[1]{>{\raggedright\arraybackslash}p{#1}}
\newcolumntype{C}[1]{>{\centering\arraybackslash}p{#1}}

\renewcommand{\arraystretch}{1.15}

\usepackage{enumitem}
\setlist{nosep}

\usepackage[
    backend=biber,
    style=apa,
    natbib=true,
    sorting=nyt,
    maxcitenames=2,
    maxbibnames=99,
    doi=true,
    url=true,
    isbn=false
]{biblatex}

\usepackage{xurl}
\usepackage[hidelinks]{hyperref}
\usepackage[nameinlink,noabbrev]{cleveref}

\title{Artificial Intelligence for Understanding and Managing
Transportation Behavior in Sustainable Smart Cities}

\author{
Junbiao Pang\\
Muhammad Ayub Sabir\\
Fatima Ashraf\\
Beijing University of Technology, Beijing, China
}

\date{}

\begin{document}

\maketitle

\begin{abstract}
Urban transportation systems generate heterogeneous data, yet these data do not automatically become actionable management intelligence. This chapter adopts a behavior-centered perspective on artificial intelligence (AI), treating mobility records and passenger-generated text as behavioral evidence rather than behavioral truth. It examines four directions: bus arrival prediction for service reliability, taxi mobility pattern discovery for demand analysis and planning, abnormal behavior detection for accountable regulatory support, and passenger-perceived risk mining for service improvement. These directions are integrated through a closed-loop framework linking data input, behavior representation, AI inference, decision support, public value, and governance feedback. The chapter identifies data quality, privacy, fairness, interpretability, uncertainty, transferability, and human accountability as essential conditions for deployment. It thereby establishes a unified pathway from behavioral evidence to operational, planning, regulatory, and passenger-service decisions.
\end{abstract}

\noindent\textbf{Keywords:} Bus Arrival Prediction, Urban Mobility Pattern Discovery, Abnormal Stop Detection, Passenger-Perceived Risk Mining, Sparse GPS, Taxi Demand, Data Governance, Human-in-the-Loop Review

\section{Introduction: AI-Enabled Transportation Behavior Management in Smart Cities}
\label{sec:introduction}

Urban transportation systems support the safe, efficient, inclusive, and low-carbon movement of people and goods. This role is central to sustainable smart cities because transportation affects service reliability, access to urban opportunities, regulatory safety, energy use, and urban economic activity. In this context, artificial intelligence (AI) should not be viewed only as a tool for improving prediction accuracy or automating isolated transportation tasks. Its broader value lies in transforming heterogeneous mobility traces into behavioral intelligence that can support operation, planning, regulation, service improvement, and sustainable urban governance \parencite{elassy2024intelligent}.

Transportation systems increasingly operate as data-rich behavioral environments. The movement of a bus, taxi, or long-distance coach records more than physical displacement; it may reflect operating decisions, passenger demand, driver routines, service conditions, and regulatory compliance. Passenger-generated digital traces, such as social media posts, further reveal perceived service risks, complaints, safety concerns, and public expectations that may not appear in conventional operational records. When these heterogeneous observations are modeled carefully, they can support a shift from passive monitoring of transportation states to proactive understanding and management of transportation behavior.

Transportation behavior remains difficult to understand because it is dynamic, heterogeneous, and often indirectly observed. Bus arrival patterns depend on route structure, stop interactions, traffic conditions, dwell-time variation, and accumulated delay propagation. Taxi pick-up and drop-off events reveal fine-grained urban mobility demand, but they appear as scattered spatial-temporal points rather than fixed-route sequences. Driver-related behaviors, such as illegal driver substitution or unauthorized stopping, are hidden within otherwise normal operation records and are often rare, imbalanced, and costly to verify. Long-distance coach abnormal stops are especially difficult to detect when GPS trajectories are sparse and abnormal labels are limited. Passenger feedback from social media is short, noisy, informal, and unevenly influenced by user engagement, yet it may contain early signals of service risk or public dissatisfaction. These characteristics show why transportation behavior cannot be managed well through isolated rules, single-source statistics, or purely manual inspection.

A central gap in many AI-enabled transportation studies is the weak connection between technical outputs and transportation governance value. Task-specific metrics such as prediction error, classification accuracy, clustering quality, and topic coherence are important, but they do not by themselves explain how AI supports dispatching, planning, compliance supervision, passenger service, carbon reduction, or urban economic vitality. For transportation management, the key question is not only whether an algorithm performs well on a dataset, but whether its output can be interpreted, trusted, deployed, and connected to decisions that improve mobility systems. This chapter therefore adopts a behavior-centered perspective: transportation data are interpreted as behavioral signals, AI methods infer predictions, patterns, anomalies, or risks from these signals, and the resulting intelligence is used to support value-oriented transportation management.

The chapter is organized around a closed-loop logic of data input, behavior representation, intelligent inference, decision support, public value, and iterative improvement. Heterogeneous data sources provide observations of movement, demand, compliance, and public perception. These observations are converted into behavior representations that preserve their transportation meaning. AI models then support different forms of intelligent inference, including operational prediction, mobility pattern discovery, abnormal behavior detection, and passenger-perceived risk mining. The inferred intelligence supports practical decisions such as dynamic scheduling, resource allocation, targeted inspection, service adjustment, and planning feedback. The outcomes of these decisions contribute to public value, including service reliability, safety, regulatory fairness, low-carbon operation, and economic vitality. Feedback from management outcomes can further improve data collection, model design, and decision processes, forming an iterative behavior-intelligence loop.

This logic also defines the main principles followed throughout the chapter. First, AI models should be behavior-centered: they should preserve the meaning of movement, demand, compliance, and perception rather than treating transportation records as generic numerical inputs. Second, AI models should be value-oriented: their outputs should be evaluated not only by technical performance, but also by their ability to support management, planning, safety, service quality, sustainability, and economic efficiency. Third, AI models should be scenario-adaptive: different transportation problems require different levels of data availability, model complexity, interpretability, computational cost, and human involvement. These principles allow the chapter to compare AI methods at a macro level, focusing on applicability, deployment conditions, data requirements, and management value rather than only marginal performance differences on isolated datasets.

A behavior-centered view also requires a trustworthy AI foundation. Transportation AI systems often involve sensitive trajectory data, operational records, driver behavior information, and public-opinion data. Their outputs may influence passenger information, resource allocation, service access, regulatory inspection, or public communication. Therefore, AI should be treated as a decision-support tool rather than an autonomous authority. In this chapter, trustworthy AI is understood through four principles: privacy-preserving data governance, fairness-aware evaluation, interpretable and accountable decision support, and human-in-the-loop final authority. These principles are consistent with broader trustworthy-AI guidance that emphasizes reliability, accountability, explainability, privacy, and fairness in deployed AI systems \parencite{ai2023artificial}. In compliance and regulatory applications, AI can help identify suspicious patterns and improve inspection efficiency, but legal and administrative judgment must remain grounded in regulations, evidence, and human review.

The chapter draws on six representative AI studies to illustrate how transportation behavior can be predicted, discovered, detected, and monitored for smart-city management. Bus arrival prediction shows how sequential learning can model operational behavior and support service reliability, passenger information, and dynamic dispatching \parencite{pang2018learning}. Taxi mobility pattern discovery shows how point-process modeling and decomposition can convert scattered trip events into interpretable urban demand structures \parencite{pang2017discovering}. Illegal driver substitution detection and abnormal coach stop detection demonstrate how routine modeling, self-similarity analysis, low-rank sparse learning, and graph-based semi-supervised learning can support compliance supervision under noisy, sparse, and imbalanced conditions \parencite{pang2024finding,deng2026unsupervised,sabir2025few}. Passenger-perceived transit risk mining from social media shows how importance-aware topic modeling can capture weak signals of service problems and public concerns that complement vehicle-based sensing \parencite{ashraf2025importance}. Table~\ref{tab:representative_studies} summarizes how these studies support the chapter's behavior-management framework.

\begin{table}[!htbp]
\centering
\caption{Representative AI studies and their roles in the chapter.}
\label{tab:representative_studies}
\scriptsize
\setlength{\tabcolsep}{3pt}
\renewcommand{\arraystretch}{1.05}
\begin{tabularx}{\textwidth}{L{2.4cm} L{2.3cm} L{2.8cm} Y Y}
\toprule
\textbf{Representative task} & \textbf{Data source} & \textbf{Core AI method} & \textbf{Behavior intelligence} & \textbf{Management value} \\
\midrule
Bus arrival prediction \parencite{pang2018learning} 
& Bus GPS and route-operation records 
& Sequential learning with heterogeneous route representation 
& Future arrival dynamics and route-progression behavior 
& Passenger information, dispatching support, service reliability, and planning feedback \\

\midrule
Taxi mobility pattern discovery \parencite{pang2017discovering} 
& Taxi pick-up and drop-off events 
& Point-process intensity modeling, decomposition, and basis-pattern factorization 
& Regular mobility structures and time-specific demand disparities 
& Hotspot monitoring, resource allocation, demand understanding, and spatial planning \\

\midrule
Illegal driver substitution detection \parencite{pang2024finding} 
& Taxi GPS, taximeter, and operation records 
& Driver-routine modeling, self-similarity analysis, and multiple-instance learning 
& Suspicious changes in rest and pick-up behavior routines 
& Targeted inspection, regulatory screening, and accountable human review \\

\midrule
Low-frequency coach abnormal stop detection \parencite{deng2026unsupervised} 
& Long-distance coach GPS trajectories 
& Stop inference with low-rank and sparse abnormal-evidence modeling 
& Candidate abnormal stop segments under low-frequency sensing 
& Cross-regional supervision, inspection planning, and safety regulation support \\

\midrule
Few-shot sparse-GPS abnormal stop detection \parencite{sabir2025few} 
& Sparse GPS trajectories with limited labels 
& Sparsity-aware segmentation, temporal indicators, graph learning, and self-training 
& Weak abnormal-stop evidence under sparse and few-shot conditions 
& Low-resource monitoring, scalable screening, and reduced inspection burden \\

\midrule
Passenger-perceived transit risk mining \parencite{ashraf2025importance} 
& Passenger-generated social media posts and metadata 
& Importance-aware topic modeling for short and noisy public feedback 
& Passenger-perceived service-risk topics and public concerns 
& Service diagnosis, complaint prioritization, public communication, and feedback response \\
\bottomrule
\end{tabularx}
\end{table}

Together, these studies show that AI-enabled transportation behavior intelligence is valuable not as a collection of isolated algorithms, but as a bridge between mobility data and actionable governance. For operations, AI can help agencies anticipate delays, adjust dispatching, and improve service reliability. For planning, AI-derived behavior patterns can reveal mobility hotspots, bottlenecks, and mismatches between supply and demand. For regulation, AI can support targeted inspection while maintaining human accountability in enforcement decisions. For passenger service, AI can detect public concerns and support faster service-improvement feedback. For sustainability and urban development, behavior intelligence can support lower-carbon mobility pathways and more efficient access to jobs, services, commercial areas, and public facilities. These contributions should be understood as pathways of management value rather than automatic outcomes; their realization depends on data quality, institutional readiness, deployment design, and trustworthy governance.

The remainder of this chapter is organized as follows. \Cref{sec:data_sources} introduces heterogeneous transportation data sources and explains how they function as behavioral signals and as the data foundation for trustworthy AI. \Cref{sec:prediction} discusses AI for transportation behavior prediction, with bus arrival prediction as a representative case of operational behavior intelligence, including macro-level algorithm selection, deployability, and emerging small--large model collaboration. \Cref{sec:mobility_patterns} presents AI methods for urban mobility pattern discovery from taxi trip records and discusses their relevance to demand understanding, resource allocation, planning support, and low-carbon mobility. \Cref{sec:abnormal_detection} examines abnormal transportation behavior detection and regulation, focusing on driver behavior monitoring, abnormal coach stop detection, sparse GPS conditions, and human-accountable enforcement support. \Cref{sec:risk_mining} discusses passenger-perceived transit risk mining from social media and its role in service improvement, public-risk perception, and fair representation of passenger concerns. \Cref{sec:integrated_framework} integrates these directions into a unified framework for transportation behavior intelligence and management. \Cref{sec:challenges} summarizes technical, ethical, and deployment challenges, including privacy, fairness, interpretability, accountability, data sparsity, model transferability, and system integration. \Cref{sec:conclusion} concludes the chapter by highlighting the role of AI-enabled behavioral intelligence in building safer, more reliable, more inclusive, and more sustainable smart-city transportation systems.

\section{Heterogeneous Data Sources for Transportation Behavior Understanding}
\label{sec:data_sources}

Heterogeneous data are the foundation of transportation behavior intelligence. AI-enabled transportation management depends not only on whether data are available, but also on whether they preserve meaningful evidence about movement, demand, compliance, service quality, and public perception. In smart cities, transportation behavior becomes observable through vehicle trajectories, operational records, trip events, driver activity logs, sparse GPS traces, and passenger-generated text. Each data source captures a different part of the transportation system, and each has different implications for modeling, deployment, privacy, and management use.

No single data modality provides a complete view of transportation behavior. Bus GPS records describe route-constrained public-transit movement; taxi pick-up and drop-off events reveal spatial-temporal travel demand; taxi operation records expose driver routines and possible compliance changes; long-distance coach trajectories indicate route-following and stopping behavior; sparse GPS data reflect low-resource monitoring conditions; and social media posts capture passenger-perceived risks and service concerns. The main task is therefore to convert raw observations into behavior representations that remain connected to transportation decisions. Without this conversion, AI models may produce outputs that are technically accurate but difficult to interpret or use in real management contexts.

Bus GPS and operational records provide direct evidence of public-transit movement behavior. They contain vehicle locations, route information, stop sequences, arrival and departure times, travel distances, and temporal operating conditions. Their primary management relevance lies in service reliability because delayed or unstable bus arrival behavior affects passenger waiting time, scheduling quality, dispatching efficiency, and public confidence in transit services. Long-term bus operation data can also support line planning, station-layout adjustment, and the identification of persistent bottlenecks. The main AI challenge is that bus arrival behavior depends on route topology, stop-level interactions, traffic conditions, current vehicle state, and accumulated movement history. This makes bus GPS data suitable for sequential and spatiotemporal models that learn how operational behavior evolves along a route \parencite{pang2018learning}.

Taxi pick-up and drop-off records represent urban mobility demand rather than fixed-route operation. Each pick-up or drop-off event reflects a passenger decision to start or end a trip at a specific location and time. When aggregated across a city, these events reveal activity centers, mobility hotspots, commuting rhythms, and time-dependent changes in travel demand. Their management value is connected to resource allocation, hotspot monitoring, and spatial planning. For operators, demand patterns can support targeted dispatching and reduce inefficient cruising. For planners, repeated trip patterns can reveal mismatches between transport supply and urban activity distribution, helping to inform infrastructure planning, commercial-area accessibility, and multimodal service coordination. Because taxi trip events appear as scattered spatial-temporal points, the AI task is to transform fragmented observations into interpretable mobility patterns \parencite{pang2017discovering}.

Taxi GPS, taximeter, and operation records support driver-routine analysis and compliance supervision. These records include movement traces, occupied and unoccupied states, pick-up behavior, rest patterns, and temporal activity characteristics. In this setting, the data do not only describe mobility demand; they also indicate whether a vehicle is being operated in a way that is consistent with the registered driver's normal routine. This is important for detecting suspicious behaviors such as illegal driver substitution. However, this data type is sensitive because it may involve driver behavior and regulatory inspection. AI should therefore be positioned as a decision-support tool: it can help identify suspicious patterns, prioritize inspection, and reduce broad manual patrols, but final regulatory judgment must remain grounded in legal procedures, operational rules, and human review. The main technical challenge is that abnormal or illegal behaviors are rare, hidden, and highly imbalanced relative to normal operations \parencite{pang2024finding}.

Long-distance coach GPS trajectories provide another compliance-oriented data source. They record locations, timestamps, speeds, engine states, and route progression along intercity corridors. These signals can reveal abnormal stopping behavior, including unauthorized passenger pick-up or drop-off at non-designated locations. Compared with urban taxi data, coach operation is more route-constrained, but the sensing condition is often more difficult. GPS records may be sparse, and a coach may stop between two consecutive samples without producing a dense trajectory trace. The behavioral evidence must therefore be inferred indirectly from speed change, distance, time interval, and road-segment stop duration. The management value of this data lies in targeted supervision, safety protection, and more efficient inspection planning, especially for cross-regional transportation systems. As with driver-compliance analysis, AI outputs should be treated as screening evidence or inspection clues rather than final enforcement decisions \parencite{deng2026unsupervised}.

Sparse GPS trajectories highlight the low-resource conditions of real transportation monitoring. In many operational systems, GPS observations are collected at relatively low frequency, and abnormal events are labeled only in small numbers. These conditions create two linked difficulties: the trajectory must first be segmented into meaningful behavioral units, and the model must then learn abnormal patterns from limited supervision. Sparse GPS data are therefore important not only as a technical challenge, but also as a deployment issue. Dense sensing and large-scale manual labeling may be unrealistic for smaller cities, intercity routes, or resource-constrained agencies. AI methods that combine domain-specific indicators, local temporal adjustment, graph relations, and semi-supervised learning can reduce the threshold for intelligent monitoring. The broader value of sparse-data modeling is inclusive deployment because it allows transportation behavior intelligence to be applied even when sensing infrastructure and labeled data are limited \parencite{sabir2025few}.

Passenger-generated social media posts extend transportation behavior understanding beyond vehicle and operation records. Posts from platforms such as Weibo may contain public feedback about delays, crowding, payment problems, service quality, safety incidents, and perceived risks. This data source captures perception behavior: what passengers notice, report, amplify, or worry about. Its value lies in complementing operational data with public experience signals. A vehicle-based system may show that a route is operating normally, while passenger comments may reveal perceived crowding, poor information service, safety concerns, or dissatisfaction with transfer conditions. Social media data are short, noisy, informal, and unevenly shaped by user engagement. Therefore, social media analysis requires AI methods that combine linguistic structure, user influence, noise control, and fairness-aware interpretation. The management goal is not only topic extraction, but also service-improvement feedback and fairer representation of passenger concerns \parencite{ashraf2025importance}.

Table~\ref{tab:data_behavior_management} summarizes the main data sources discussed in this chapter. The table is organized around behavioral evidence, AI role, management use, governance concern, and core value. This structure keeps the focus on transportation behavior management rather than presenting the data sources as isolated technical inputs.

\begin{table}[!htbp]
\centering
\caption{Heterogeneous transportation data sources and their behavioral-management roles.}
\label{tab:data_behavior_management}
\scriptsize
\setlength{\tabcolsep}{3pt}
\renewcommand{\arraystretch}{1.05}
\begin{tabularx}{\textwidth}{L{2.3cm} Y L{2.6cm} Y Y L{2.3cm}}
\toprule
\textbf{Data source} & \textbf{Behavioral evidence} & \textbf{AI role} & \textbf{Management use} & \textbf{Data and governance concern} & \textbf{Core value} \\
\midrule
Bus GPS and operational records 
& Route progression, stop-to-stop travel, arrival dynamics, and service regularity 
& Operational behavior prediction 
& Passenger information, dynamic dispatching, delay monitoring, and planning feedback 
& Structured operator data; real-time use requires stable interfaces and route-level validation 
& Reliability, operational efficiency, and planning support \\

\midrule
Taxi pick-up and drop-off records
& Spatial-temporal demand, hotspots, activity centers, and trip-end patterns
& Mobility pattern discovery
& Resource allocation, hotspot monitoring, station-area analysis, and spatial planning
& Trip-level platform or operator data; aggregation is needed to reduce privacy and sparsity risks
& Demand matching, reduced cruising, and urban vitality \\

\midrule
Taxi GPS, taximeter, and operation records
& Driver routines, operating states, pick-up behavior, rest patterns, and routine changes
& Suspicious-operation analysis
& Compliance supervision, inspection prioritization, and driver-behavior review
& Sensitive operational logs; access control, audit trails, and human review are necessary
& Regulatory efficiency, safety, fairness, and privacy-aware supervision \\

\midrule
Long-distance coach GPS trajectories
& Route-following behavior, speed change, stop evidence, and abnormal stop patterns
& Abnormal stop inference
& Cross-regional supervision, candidate stop-spot discovery, and targeted inspection
& Often sparse and cross-jurisdictional; interpretation requires road context and regulatory rules
& Safety regulation and accountable enforcement support \\

\midrule
Sparse GPS trajectories
& Incomplete movement traces, weak stop evidence, local temporal changes, and limited abnormal labels
& Few-shot or semi-supervised abnormal behavior recognition
& Low-resource monitoring, scalable screening, and weak-evidence expansion
& Limited observability and label scarcity; pseudo-labels require careful control and verification
& Inclusive deployment and reduced monitoring cost \\

\midrule
Passenger-generated social media posts
& Complaints, perceived risks, service concerns, emotional responses, and passenger experience
& Passenger-perceived risk mining
& Risk awareness, service diagnosis, complaint prioritization, and public-feedback analysis
& Noisy and biased unstructured text; active users may overrepresent public attention
& Passenger experience, public trust, and fair representation \\
\bottomrule
\end{tabularx}
\end{table}

\subsection{Trustworthy Data Governance for Transportation Behavior Intelligence}

Trustworthy transportation AI begins with trustworthy data governance. Because transportation data may involve movement traces, driver activity, operational decisions, and public-opinion records, data use must be guided by applicable laws, regulations, institutional rules, and public-interest principles. Technology should support this governance process, but it should not replace it. In particular, data-processing technologies should help improve privacy protection, consistency, traceability, and compliance; they should not create independent regulatory standards or make final administrative judgments. This view is consistent with risk-management guidance that treats trustworthy AI as a socio-technical issue involving data, models, people, organizations, and deployment context \parencite{ai2023artificial}.

A first principle is privacy minimization. Data collection and processing should be limited to what is necessary for the transportation-management purpose. Trajectory data, taximeter records, driver-routine information, and passenger-generated posts should be aggregated, anonymized, or otherwise protected whenever individual-level identification is not required. Privacy-preserving technologies, such as aggregation, access control, differential privacy, and federated learning, can support safer use of transportation data, but their role is auxiliary: they help implement governance requirements rather than define those requirements independently.

A second principle is compliant cross-domain collaboration. Transportation behavior management often requires data from multiple actors, such as transit operators, taxi platforms, coach companies, road authorities, and public-service departments. These sources may be controlled by different institutions and governed by different access rules. Federated or distributed learning frameworks may help reduce the need to move raw data across organizational boundaries, which is why federated learning has attracted increasing attention in intelligent transportation systems. However, federated learning also introduces practical challenges, including uneven data distributions, limited computing resources, communication constraints, and remaining privacy or security risks \parencite{zhang2023federated}. Therefore, such methods should be deployed only within clear legal and institutional agreements that define data ownership, permitted use, security responsibility, and audit procedures.

A third principle is data fairness. Transportation data do not represent all groups and regions equally. Central urban areas may have denser sensing infrastructure, while suburban or low-resource areas may produce fewer digital traces. Social media data may overrepresent highly active users and underrepresent passengers who rarely post online. If such biases are ignored, AI systems may produce unequal service recommendations, uneven prediction quality, or biased risk perception. Data governance should therefore include coverage assessment, bias diagnosis, and fairness-aware sampling or calibration where appropriate.

A fourth principle is traceable quality management. AI decisions in transportation management should be supported by data whose origin, processing steps, and quality limitations can be inspected. This is especially important for regulatory or service-critical applications. If a model flags a suspicious stop, predicts a major delay, or identifies a public-risk topic, managers should be able to trace the underlying data source and understand whether the output may be affected by missing data, GPS noise, platform bias, preprocessing decisions, or label uncertainty. Traceability strengthens both technical reliability and institutional accountability.

Several cross-cutting challenges remain. Transportation data are heterogeneous, combining spatial points, temporal sequences, trip records, categorical states, graph structures, and textual posts. Many behavioral signals are sparse or incomplete, requiring methods that can learn from weak evidence and limited labels. Transportation behavior is also imbalanced: normal operations dominate the data, while risky or illegal events are rare but important. Observations are noisy because of GPS drift, incomplete logs, duplicated posts, slang, and inconsistent human behavior. Finally, behavioral interpretation requires domain knowledge. A stop, delay, crowded station, or driver-routine change becomes meaningful only when connected to route structure, operating rules, passenger demand, service standards, and regulatory context.

These challenges explain why AI for transportation behavior management must be both data-aware and governance-aware. Data should not be treated as generic inputs for model training. They should be interpreted as behavioral evidence, processed under trustworthy governance principles, and connected to specific management purposes. This data foundation prepares the next sections, where transportation behavior evidence is transformed into prediction, mobility pattern discovery, abnormal behavior detection, and passenger-perceived risk mining.

\section{AI for Transportation Behavior Prediction}
\label{sec:prediction}

Transportation behavior prediction estimates future operational states from observed mobility histories. In public-transit systems, bus arrival time prediction is a representative task because it directly connects vehicle movement, passenger waiting time, dispatching efficiency, service reliability, and planning feedback. This section uses bus arrival prediction to show how AI can transform sequential operation records into practical behavioral intelligence for public-transit management.

The task is challenging because buses operate in a dynamic environment. Although each bus follows a planned route and stop sequence, its actual arrival behavior is affected by traffic conditions, stop dwell time, intersections, passenger activity, route topology, and the accumulated influence of previous segments. A delay at one stop may propagate to several downstream stops, while a short local speed fluctuation may have only limited impact. Effective prediction therefore requires models that can capture both the current vehicle state and longer temporal dependencies along the route.

\subsection{Bus Arrival Prediction as a Multi-Step Operational Behavior Problem}

The core problem in bus arrival prediction is not simply how to minimize next-stop prediction error. A management-oriented prediction system must provide reliable multi-step predictions for downstream stops under changing traffic and operational conditions. Passengers often need arrival information for upcoming stops, while operators need to anticipate route progression before delays fully propagate. Therefore, prediction output should support real-time passenger information, dynamic dispatching, service monitoring, and planning analysis.

Many conventional prediction strategies estimate the next arrival state and then recursively extend the prediction to later stops. This one-step-ahead strategy is easy to implement, but it becomes less reliable as the prediction horizon increases because early errors may propagate into later estimates. It also provides limited support for multi-stop management decisions, where operators need to understand how current conditions may affect several downstream locations. A more suitable formulation treats bus arrival prediction as a multi-step-ahead task, where the model directly learns the relationship between the current movement context and multiple future arrivals.

This framing is central to the representative bus-arrival study used in this chapter. The study emphasizes that practical bus arrival prediction often requires multi-step-ahead prediction because a current bus state must be used to estimate arrival times at several downstream stops \parencite{pang2018learning}. This view is important for transportation management: the value of prediction is not only to estimate the next stop, but also to understand how current operation will affect the route ahead.

The behavioral signal in this task is heterogeneous. GPS positions describe vehicle movement; stop sequences define route progression; distances between route points encode spatial structure; departure and current times represent temporal state; and static route information provides operational context. These signals jointly describe how a bus behaves along a route. AI prediction becomes meaningful when it preserves this heterogeneous behavioral context instead of reducing the task to a generic travel-time regression problem.

A further challenge is abnormal or unexpected operating conditions. Accidents, road construction, severe weather, large events, temporary traffic controls, and sudden demand changes may alter arrival behavior in ways that are not fully represented in historical trajectory data. Models trained only on regular GPS patterns may perform well in ordinary conditions but become less reliable under such out-of-distribution events. This motivates prediction systems that remain lightweight for routine operation while allowing external context to be considered when abnormal conditions occur.

\subsection{Heterogeneous Route Representation and Sequential Learning}

Sequential learning models are suitable for bus arrival prediction because bus movement is temporally ordered and depends on previous route states. Recurrent neural networks, especially Long Short-Term Memory (LSTM) and Gated Recurrent Unit (GRU) models, can maintain internal states that capture short-term movement variation and longer-range route dependency. This property is useful in public-transit prediction because future arrivals depend not only on the current GPS point, but also on accumulated movement history across previous stops and road segments.

A compact formulation of the prediction task is
\begin{equation}
\hat{t}_{k,k+\Delta}
=
f\left(\text{trip}, \text{network}, \text{static}, \text{other}, p_k, t_k\right),
\label{eq:bus_arrival_prediction}
\end{equation}
where $\hat{t}_{k,k+\Delta}$ denotes the predicted arrival time from the current route position $k$ to a future stop $k+\Delta$, $p_k$ and $t_k$ denote the current position and time, and $f(\cdot)$ learns from trip history, network structure, static route features, and other operational measurements \parencite{pang2018learning}. This formulation shows that bus arrival prediction depends on a combined behavioral context rather than isolated GPS coordinates.

The representative study provides useful empirical grounding for this formulation. It uses Beijing bus operation data collected from 47 routes and 1,089 buses during February 2015, with GPS observations recorded at an average frequency of approximately one point every 30 seconds \parencite{pang2018learning}. The dataset includes dynamic GPS trajectories and static route information such as bus stops, intersections, route directions, and stop-related statistics. This setting reflects real public-transit operation because arrival behavior is shaped by both vehicle movement history and route infrastructure.

The route representation should preserve both movement dynamics and spatial meaning. GPS trajectories provide dynamic vehicle states, while stops, intersections, and interpolated route points create a structured sequence of meaningful locations. Stops represent passenger service points, intersections capture possible delay sources, and interpolated points improve spatial resolution along longer road segments. In the representative LSTM-RNN study, these heterogeneous elements are encoded into a unified vector space so that the model can learn how movement evolves across service locations and road-network structures \parencite{pang2018learning}. This means the model is not only learning time values, but also learning how route context shapes operational behavior.

The same study also shows why long-range dependencies matter. In multi-step-ahead prediction, downstream arrivals are influenced by multiple previous route states. A model that only uses short dependencies may miss how earlier delays, stop interactions, or route-segment conditions affect future stops. The LSTM structure addresses this issue through memory states and gates that allow the model to preserve useful information across multiple time steps \parencite{pang2018learning}. In practical terms, the model uses observed route history to correct and stabilize predictions for future stops.

Empirical evidence from the representative study supports two methodological lessons. First, multi-step-ahead prediction benefits from long-range temporal dependencies because downstream arrivals are influenced by accumulated route history. Second, heterogeneous measurements matter because stops, intersections, and interpolated points encode different physical and operational properties of the route. The study reports that LSTM-RNN performs better than several regression-, filtering-, and searching-based baselines, and its ablation analysis shows that richer route measurements improve prediction quality \parencite{pang2018learning}. These results are used here as evidence for behavior-aware representation rather than as a narrow performance ranking.

\subsection{Prediction Routes for Deployment-Oriented System Design}

For management-oriented prediction, algorithm selection should depend on operating conditions, data availability, computational resources, and the need for interpretability. Table~\ref{tab:bus_prediction_methods} summarizes major prediction routes from a deployment-oriented perspective. The comparison focuses on scenario fit, data and deployment needs, strengths, and limitations rather than small performance differences on a single benchmark.

\begin{table}[!htbp]
\centering
\caption{Macro-level comparison of prediction routes for bus arrival prediction.}
\label{tab:bus_prediction_methods}
\scriptsize
\setlength{\tabcolsep}{3pt}
\renewcommand{\arraystretch}{1.05}
\begin{tabularx}{\textwidth}{L{2.4cm} L{2.2cm} Y Y Y Y}
\toprule
\textbf{Prediction route} & \textbf{Typical models} & \textbf{Suitable conditions} & \textbf{Data and deployment needs} & \textbf{Main strengths} & \textbf{Main limitations} \\
\midrule
Statistical and filtering route
& ARIMA, Kalman filter
& Stable routes with regular travel-time patterns and limited disruptions
& Small historical datasets; very low computing cost; simple system integration
& Interpretable, lightweight, and suitable for basic deployment
& Limited ability to model nonlinear dynamics, long-range dependencies, and abnormal events \\

\midrule
Regression and searching route
& SVR, k-NN, kernel regression
& Routes where similar historical trips or engineered features are informative
& Historical trips and handcrafted features; search-based methods may become costly at large scale
& Useful baselines; can exploit historical similarity or nonlinear feature relations
& Weak multi-step representation; may struggle when current conditions differ from historical patterns \\

\midrule
Sequential deep-learning route
& LSTM, GRU
& Urban bus routes with continuous GPS sequences and sequential dependency
& Moderate historical trajectory data; low to medium computing cost; suitable for real-time inference
& Captures temporal dependencies and supports multi-step-ahead prediction
& Limited representation of explicit road-network topology and cross-route interactions \\

\midrule
Spatiotemporal graph route
& STGNN, STGCN, ST-GAT
& Dense urban networks where stops, road segments, and routes interact
& Road-network structure plus trajectory data; medium computing cost; usually requires stronger system integration
& Models spatial and temporal dependencies jointly
& Higher deployment complexity and weaker interpretability if not carefully designed \\

\midrule
Physics-guided route
& PINN, physics-guided neural models
& Sparse or partially missing data when traffic-flow prior knowledge is available
& Domain knowledge and model-design expertise; medium computing cost
& Can improve robustness by incorporating traffic-flow constraints
& Requires reliable assumptions and careful calibration \\

\midrule
Small--large model hybrid route
& Lightweight predictor plus large-model event interpreter
& Routes affected by accidents, construction, weather, or event-driven disruptions
& Real-time GPS for the small model; external event data and cloud support for semantic interpretation
& Balances low-latency routine prediction with contextual correction under abnormal scenarios
& Requires reliable event sources, traceable correction logic, and human oversight \\
\bottomrule
\end{tabularx}
\end{table}

The first three prediction routes are closely related to the methodological landscape discussed in the representative bus-arrival study. Statistical and filtering methods are attractive when data are limited and operating patterns are stable. Regression and searching methods provide useful baselines by learning feature relations or comparing current trips with historical ones. Sequential deep-learning models are suitable when route histories are available and multi-step dependency is important. They are also practical for real-time systems because inference can remain relatively lightweight.

The remaining routes extend the discussion toward current and future deployment. Spatiotemporal graph models become useful when road-network topology, stop relations, and route interactions strongly affect arrival behavior. Recent surveys show that spatiotemporal graph neural networks are designed to model spatial and temporal dependencies jointly in urban predictive learning tasks \parencite{jin2023spatio}. Physics-guided models are promising when pure data-driven learning is unstable under sparse or incomplete sensing, because traffic-flow constraints can guide learning beyond observed samples. These methods may be useful when a city wants prediction models that are more transferable, robust, or consistent with known transportation dynamics.

Large-model-assisted prediction is a more recent direction, but it should be used carefully. Large language models and multimodal foundation models are not necessarily suitable for high-frequency real-time bus arrival prediction by themselves because they can be computationally expensive and may not directly process continuous trajectory streams efficiently. Their value is stronger in semantic understanding of external abnormal events. Road construction notices, accident reports, severe-weather warnings, traffic-control announcements, and public reports may explain why a route suddenly deviates from its usual arrival pattern. Recent work on explainable traffic prediction with large language models suggests that language-based models can help connect traffic patterns with interpretable contextual explanations \parencite{guo2024towards}. In this chapter, however, such models are treated as contextual support tools, not replacements for lightweight real-time predictors.

A practical architecture is therefore small--large model collaboration. A lightweight model, such as an LSTM, GRU, or compact spatiotemporal graph model, performs high-frequency real-time prediction under regular operating conditions. A larger semantic model operates at a lower frequency to interpret external event information and estimate its likely impact on route segments, delay propagation, or prediction uncertainty. The large model does not replace the small predictor; instead, it provides correction signals or contextual alerts when abnormal conditions occur. This design can reduce computational burden while improving robustness under unusual scenarios. It also fits transportation deployment because routine prediction remains efficient, while semantic reasoning is activated only when additional context is needed.

Trustworthy AI is essential in prediction scenarios. First, prediction outputs should be interpretable enough for dispatchers to understand whether a delay is likely caused by congestion, stop dwell time, route disruption, or abnormal external events. Second, prediction quality should not be concentrated only on central or data-rich routes; suburban or lower-frequency routes also require reliable passenger information. Third, both small-model predictions and large-model correction suggestions should remain decision-support outputs. Final dispatching decisions should be made by human operators who can consider operational constraints, passenger needs, safety rules, and institutional responsibilities.

\subsection{Management Value and Deployment Trends}

The output of AI-based bus arrival prediction is a form of operational behavior intelligence. Predicted arrival times summarize how the current trajectory, route structure, and temporal context are likely to evolve. This intelligence supports both passenger-facing and agency-facing decisions. For passengers, accurate arrival information reduces uncertainty and improves perceived service quality. For operators, multi-step prediction supports dynamic dispatching, interval adjustment, delay monitoring, and emergency response.

From a management perspective, bus arrival prediction contributes to service reliability in three main ways. First, it improves real-time passenger information by estimating when vehicles will reach future stops. Second, it supports operational control by identifying unstable or delayed route progression before the delay fully propagates. Third, it provides a data-driven basis for evaluating route performance, stop-level bottlenecks, and the effectiveness of service adjustments. These functions connect prediction directly with public-transit management rather than treating it as an isolated machine-learning task.

The planning value of bus arrival prediction emerges when prediction outputs are accumulated over longer periods. Repeated delay patterns can reveal persistent bottlenecks at particular stops, intersections, or route segments. Such evidence can support bus-lane planning, stop relocation, signal-priority design, route adjustment, and service-frequency revision. In this sense, arrival prediction is not only a real-time information service; it can also become a feedback mechanism for public-transit planning and infrastructure improvement.

The sustainability and economic values should be understood as management pathways rather than automatic effects. More reliable prediction can support better dispatching, reduce unnecessary waiting or idling, improve service regularity, and make public transit more attractive to passengers. These changes may contribute to lower energy waste and reduced reliance on private vehicles when combined with appropriate operational policies. Reliable public-transit information also reduces passenger time uncertainty, improves perceived service quality, and supports access to employment, education, commercial areas, and public services. These contributions show why prediction should be evaluated not only by model error, but also by its ability to support broader transportation-management outcomes.

Practical deployment still faces several challenges. Prediction models must generalize under accidents, weather changes, road construction, and major events. They must also transfer across routes and cities with different stop spacing, passenger demand, signal settings, and congestion patterns. A technically accurate model may still fail to deliver value if it cannot be integrated with dispatching platforms, passenger information systems, vehicle-location systems, and human operating procedures. Future systems should therefore become more adaptive, more multimodal, and more closely linked with operation and planning. This direction reflects the broader goal of AI-enabled transportation behavior management: not only to predict future states, but to support reliable, trustworthy, and sustainable mobility decisions.

\section{AI for Urban Mobility Pattern Discovery}
\label{sec:mobility_patterns}

Urban mobility pattern discovery aims to reveal how travel demand is distributed across space and time. Unlike bus arrival prediction, which models ordered movement along fixed public-transit routes, mobility pattern discovery focuses on flexible and scattered travel behavior. Taxi pick-up and drop-off records are a representative source for this task because each trip event reflects a passenger decision to start or end a journey at a specific location and time. When aggregated across a city, these events provide a fine-grained view of mobility demand, activity centers, transport hubs, commercial districts, residential areas, and time-varying travel needs.

The value of this task is not limited to producing heat maps. For transportation management, urban mobility patterns can support taxi resource allocation, hotspot monitoring, event response, multimodal service coordination, and spatial planning. They can also provide evidence for reducing inefficient cruising by helping vehicles move toward likely demand areas. From a broader urban-governance perspective, taxi trip patterns act as behavioral signals of how people access jobs, services, stations, shopping areas, hospitals, schools, and entertainment districts. AI is therefore needed to transform scattered trip points into interpretable mobility intelligence that can support both operational decisions and long-term planning.

\subsection{Taxi Trip Events as Urban Demand Signals}

Taxi pick-up and drop-off events are useful because they represent revealed mobility demand. A pick-up point indicates where a passenger requests transport service, while a drop-off point indicates where a trip ends and where urban activity may be taking place. These two signals are related but not identical. Pick-up patterns may reflect residential origins, transport hubs, or departure points from activity areas, while drop-off patterns may reveal destinations, employment centers, entertainment zones, or time-sensitive urban functions.

However, taxi trip data are difficult to interpret directly. At a fine spatial scale, trip points are often sparse. A small grid cell may contain very few pick-up or drop-off events within a short time interval, even in a large city. Direct counting may therefore produce unstable and fragmented patterns. At the same time, taxi demand is not purely random. It contains repeated regularities, such as commuting rhythms and persistent station-area demand, but it also contains time-specific disparities, such as event-related hotspots, late-night entertainment activity, temporary congestion effects, or sudden demand shifts.

This creates the core behavioral problem of urban mobility pattern discovery: how to distinguish stable mobility structures from time-specific deviations when the raw observations are sparse and spatially fragmented. Regular patterns are important for strategic planning and long-term service design, while disparities are important for short-term dispatching, hotspot supervision, and event response. A useful AI method must therefore preserve both dimensions.

The representative taxi-pattern study addresses this problem using large-scale Beijing taxi data. The dataset was generated by approximately 53,000 taxis over one week, from August 15 to August 21, 2016, within the Fifth Ring Road area. After extracting meter-transition events, the study identified 4,011,302 pick-up points and 3,967,140 drop-off points \parencite{pang2017discovering}. This scale shows that taxi trip records can act as a city-scale behavioral sensor. Yet even with millions of trip events, fine-grained analysis remains difficult because the data become sparse when divided into small spatial and temporal units.

\subsection{From Point Intensities to Regularity and Disparity}

A key idea in fine-grained mobility pattern discovery is to avoid relying only on raw point counts. Instead, scattered pick-up and drop-off events can first be converted into spatial intensity surfaces. The representative study uses a Log-Gaussian Cox Process (LGCP) to estimate point intensities over space, allowing smoother demand surfaces to be inferred even when observed points are sparse \parencite{pang2017discovering}. In this way, taxi trip events are transformed from isolated points into a continuous representation of mobility intensity.

After intensity estimation, the next step is to separate repeated mobility regularity from time-specific disparity. A compact formulation of this idea is
\begin{equation}
\lambda_t = B + H_t = B + \sum_{k=1}^{K} c_{t,k}p_k ,
\label{eq:mobility_decomposition}
\end{equation}
where $\lambda_t$ denotes the estimated mobility intensity at time slice $t$, $B$ represents the repeated regularity shared across time slices, and $H_t$ represents the time-specific disparity. The disparity component is further factorized into spatial basis patterns $p_k$ with nonnegative temporal weights $c_{t,k}$ \parencite{pang2017discovering}. This equation is useful for the chapter because it gives a clear behavioral interpretation: observed mobility intensity is composed of persistent urban structure and temporary demand variation.

The low-rank regularity component captures patterns that recur across time. These may correspond to railway stations, airports, intercity bus stations, central business districts, residential clusters, or major corridors that repeatedly generate travel demand. The sparse disparity component captures deviations that are not explained by the regular structure. These may correspond to event-related activity, unusual hotspots, late-night demand, or localized imbalance between passenger demand and vehicle supply.

Nonnegative matrix factorization further improves interpretability by expressing the disparity component through a small number of basis patterns and temporal weights. Each basis pattern can be understood as an interpretable spatial activity pattern, while the corresponding weight indicates how strongly that pattern appears in a specific time slice. This makes it possible to compare different time periods, such as morning peak, evening peak, late night, or weekend activity, using a common set of spatial patterns.

The technical pipeline can therefore be understood as a behavior-interpretation process rather than only an algorithmic process. Table~\ref{tab:taxi_pattern_pipeline} shows how each stage transforms taxi trip events into management-relevant mobility intelligence.

\begin{table}[!htbp]
\centering
\caption{Behavioral interpretation pipeline for taxi mobility pattern discovery.}
\label{tab:taxi_pattern_pipeline}
\scriptsize
\setlength{\tabcolsep}{3pt}
\renewcommand{\arraystretch}{1.05}
\begin{tabularx}{\textwidth}{L{2.5cm} Y Y Y}
\toprule
\textbf{Stage} & \textbf{Technical operation} & \textbf{Behavioral meaning} & \textbf{Management use} \\
\midrule
Taxi trip events
& Extract pick-up and drop-off locations from taxi operation records
& Revealed passenger demand and urban activity signals
& City-scale demand sensing and service monitoring \\

\midrule
Point intensity estimation
& Use LGCP to estimate smooth spatial intensity from sparse point observations
& Fine-grained demand surface under sparse observations
& Hotspot monitoring and spatial demand visualization \\

\midrule
Regularity extraction
& Decompose mobility intensity into a low-rank repeated component
& Persistent mobility structures shared across time slices
& Long-term planning, hub analysis, and service-design support \\

\midrule
Disparity extraction
& Separate sparse time-specific deviations from regular demand
& Temporary hotspots, special events, or short-term demand imbalance
& Adaptive dispatching, event response, and targeted supervision \\

\midrule
Basis-pattern factorization
& Factorize disparity into nonnegative spatial basis patterns and temporal weights
& Comparable activity patterns across different time slices
& Peak-period analysis, activity-zone interpretation, and resource allocation \\

\midrule
Domain interpretation
& Link patterns with stations, business districts, roads, land use, and local knowledge
& Urban functional meaning behind mobility patterns
& Planning feedback, policy discussion, and inspection-resource prioritization \\
\bottomrule
\end{tabularx}
\end{table}

This pipeline is valuable because it connects computation with interpretation. The model does not only detect where taxi trips are dense. It explains which parts of the city show repeated demand, which parts become active only at certain times, and how these patterns relate to urban functions. In the representative study, discovered regularities and basis patterns corresponded to physically meaningful places such as railway stations, airports, intercity coach stations, business areas, and other activity zones \parencite{pang2017discovering}. Such interpretation is essential for transportation managers because a pattern becomes useful only when it can be connected to a place, a time, a service need, or a planning question.

\subsection{Mobility Pattern Intelligence for Planning and Management}

The output of urban mobility pattern discovery is a set of interpretable spatial-temporal demand structures. These structures can support both operational management and long-term planning. For taxi operations, discovered hotspots can guide vehicle allocation and reduce inefficient cruising. For traffic management, repeated activity areas can indicate where curbside management, taxi stands, or passenger loading zones may be needed. For multimodal planning, taxi demand around rail stations, bus terminals, airports, and commercial districts can reveal how taxi services interact with other transport modes.

Regularity patterns are especially useful for strategic planning. They identify locations where demand appears repeatedly across many time slices. Such locations may require stable service supply, infrastructure support, better connection to public transport, or improved traffic organization. In the taxi-pattern study, the regularity component revealed meaningful urban activity structures, showing that city mobility follows repeated spatial-temporal rules rather than purely random point distributions \parencite{pang2017discovering}. For planners, this type of evidence can support decisions about mobility hubs, road-space allocation, station-area management, and service coverage.

Disparity patterns are useful for adaptive management. They identify places and times where demand deviates from the regular structure. Such deviations may indicate event-related activity, temporary imbalance, late-night travel needs, or sudden changes in service demand. Because the disparity component is factorized into basis patterns and weights, managers can compare different time slices and understand which spatial pattern dominates at a given time. This supports more flexible dispatching and targeted response.

The paper also provides a direct example of management use: discovered patterns can help allocate limited law-enforcement or inspection resources to areas where taxi activity is persistent and intensive \parencite{pang2017discovering}. In a book-chapter context, this example should be understood as decision support rather than automatic enforcement. AI-derived mobility patterns can help identify where inspection or service management may be most needed, but any regulatory action should remain guided by institutional rules and human judgment.

The low-carbon value of mobility pattern discovery should be framed as a management pathway. If demand hotspots and temporal patterns are better understood, vehicles can be guided more efficiently, passenger waiting time can be reduced, and supply-demand mismatch can be mitigated. These improvements may reduce unnecessary cruising and energy waste when combined with appropriate dispatching policies or platform-level guidance. The chapter therefore treats mobility pattern discovery as part of a broader behavior-intelligence chain: taxi trip events reveal demand; AI transforms them into interpretable patterns; patterns support allocation and planning; better allocation can contribute to more efficient and potentially lower-carbon urban mobility.

The economic value is also important. Better demand understanding can improve driver productivity, reduce wasted operating time, and improve passenger access to mobility services. At the city level, mobility patterns can help planners understand the relationship between transportation activity and commercial areas, employment centers, tourism sites, and residential communities. Such evidence can support urban economic vitality by improving access to active areas and identifying locations where transport supply constrains activity.

\subsection{Deployment Considerations and Future Directions}

Urban mobility pattern discovery faces several deployment challenges. The first is data access and representativeness. Taxi trip data are often controlled by operators, platforms, or public agencies, and they may not represent all forms of urban travel. If only taxi data are analyzed, the resulting patterns may reflect a partial view of mobility demand. Future systems should combine taxi data with public transit, ride-hailing, shared mobility, walking, cycling, land-use, and event data where governance conditions allow.

The second challenge is spatial and temporal resolution. Fine-grained analysis can reveal local patterns, but it also increases sparsity and privacy risk. Coarse analysis is easier to deploy and communicate, but it may hide important local variation. Future methods should use adaptive spatial-temporal resolution, where the level of detail is selected according to data density, management purpose, and privacy constraints.

The third challenge is interpretation. Mobility patterns are useful only when they can be connected to urban functions and management actions. A discovered basis pattern or cluster should be interpreted in relation to transport hubs, residential areas, business districts, schools, hospitals, tourism sites, events, or land-use characteristics. This requires collaboration between AI models and transportation-domain knowledge. Semantic enrichment with points of interest, road-network features, event calendars, and planning data can improve interpretation, but such external information must be verified and used carefully.

The fourth challenge is fairness and governance. Data-driven dispatching or hotspot prioritization may unintentionally concentrate service in high-demand or high-income areas while reducing attention to peripheral or low-demand communities. Mobility intelligence should therefore be used to improve system-wide accessibility, not only to maximize short-term demand capture. Fairness-aware evaluation is needed when mobility patterns are used for resource allocation, planning, or service recommendations.

Recent trajectory-computing research also shows that mobility-data mining is moving beyond single-source pattern extraction toward broader systems that include preprocessing, representation learning, forecasting, anomaly detection, recommendation, mobility generation, and large-model-assisted interpretation \parencite{chen2024trajectory}. For transportation management, this means future mobility pattern discovery should become more multimodal, more interpretable, and more closely connected to decision workflows.

Future development will likely move in three directions. First, mobility pattern discovery will become more multimodal by integrating taxi trips with public transit, ride-hailing, shared mobility, land-use, and event data. Second, models will become more interpretable and semantically grounded, linking discovered patterns to urban functions rather than presenting only abstract clusters. Third, mobility intelligence will be more closely connected to operational and planning decisions, forming a feedback loop from demand observation to resource allocation, planning adjustment, and policy evaluation. This direction is consistent with the broader goal of AI-enabled transportation behavior management: to convert scattered mobility traces into trustworthy and actionable knowledge for sustainable smart-city governance.

\section{AI for Abnormal Transportation Behavior Detection and Regulation}
\label{sec:abnormal_detection}

Abnormal transportation behavior detection focuses on identifying operational patterns that may indicate safety risks, service violations, or regulatory non-compliance. Compared with prediction and mobility pattern discovery, this task is more sensitive because model outputs may influence inspection decisions and regulatory attention. Therefore, AI should be positioned as a decision-support tool. It can flag suspicious patterns, prioritize inspection resources, and provide interpretable evidence for human review, but it should not directly determine legal guilt, punishment, or administrative responsibility.

This governance boundary is central to trustworthy transportation AI. Vehicle trajectories, taximeter records, driver routines, and coach GPS traces may contain sensitive information about drivers, operators, passengers, and companies. A suspicious model output may reflect a true violation, but it may also arise from data noise, emergency stopping, special dispatching arrangements, temporary route conditions, incomplete records, or legitimate operational variation. For this reason, regulatory AI should support evidence organization and inspection prioritization, while final judgment must remain grounded in regulations, field verification, administrative procedures, and human authority.

This section discusses three representative abnormal-behavior scenarios: illegal driver substitution in taxi services, unauthorized long-distance coach stopping, and sparse-GPS abnormal stop monitoring under limited labels. Together, these scenarios show how AI can transform noisy and incomplete operational data into regulatory clues while maintaining a clear boundary between computational screening and legal determination.

\subsection{Abnormal Operation Detection as Evidence-Oriented Screening}

The core problem in abnormal transportation behavior detection is that many violations are rare, hidden, and costly to inspect manually. Illegal driver substitution may occur when a taxi is operated by someone other than the legally registered driver. Unauthorized coach stopping may occur when a long-distance coach stops at a non-designated location to pick up or drop off passengers. Both behaviors are difficult to observe directly because they are embedded within otherwise normal transportation operations.

Manual inspection alone is often inefficient. In taxi regulation, the number of vehicles is large, while the number of law-enforcement officers is limited. Inspectors may have to rely on experience, complaints, or random checks. In coach supervision, unauthorized stops may occur along long intercity routes, often outside fixed inspection stations. Drivers may avoid routine checkpoints or arrange informal pick-ups at convenient roadside locations. These conditions create a mismatch between regulatory workload and inspection capacity.

AI can help reduce this mismatch by detecting behavior patterns that deserve further review. However, abnormal behavior should not be interpreted as confirmed violation. A taxi may show changed patterns because of maintenance, legitimate shift adjustment, special demand, or temporary operating conditions. A coach may stop because of congestion, traffic signals, severe weather, passenger illness, or safety concerns. Therefore, the output of AI should be treated as an inspection clue or risk-prioritization result, not as a final regulatory decision.

This framing changes the technical objective. The goal is not to build a model that ``judges'' a driver or operator. The goal is to construct a traceable evidence chain: operational data are converted into behavior representations, the model identifies suspicious deviations, and human inspectors review the case using legal and contextual evidence. This evidence-oriented logic is especially important for fairness, accountability, and public trust.

\subsection{Driver-Routine Modeling for Illegal Driver Substitution Screening}

Illegal driver substitution detection is based on the idea that registered taxi drivers often have relatively stable behavior routines. Taxi GPS traces and taximeter records can reveal resting locations, rest durations, working periods, pick-up behavior, occupied and vacant states, income-related service records, and spatial-temporal passenger-search strategies. When a vehicle is operated by an unauthorized person, these routine patterns may change.

The representative IDS study formulates the problem as a supervised screening task. Given a taxi $c$, GPS-based feature extractor $\phi_T(\cdot)$, and taximeter-record feature extractor $\phi_R(\cdot)$, the model learns
\begin{equation}
f\left(\phi_T(c), \phi_R(c)\right) \rightarrow \{+1,-1\},
\label{eq:ids_screening}
\end{equation}
where $+1$ denotes a taxi suspected of IDS activity and $-1$ denotes a taxi without such evidence \parencite{pang2024finding}. In this chapter, this formulation is interpreted as a screening model: it ranks or flags taxis for inspection, rather than making a final legal conclusion.

A key lesson from the IDS study is that static driver profiles are unreliable for this task. Attributes such as age, education level, or the first time of obtaining a vocational license do not provide strong separation between suspicious taxis and normal role-model taxis. The study also reports that IDS cases are rare, with only about 0.19\% of taxis in the focused Beijing study implicated in IDS activity from January 2015 to September 2016 \parencite{pang2024finding}. This rarity makes broad manual inspection inefficient and motivates behavior-based screening.

The study focuses on two behavior sources: Sleeping Time and Location (STL) behavior and Pick-Up (PU) behavior. STL behavior captures rest-related patterns. For one-shift taxis, a registered driver often has relatively stable rest locations and rest durations because a person usually has a relatively fixed residence or rest routine. If another person operates the taxi, the sleep-location and rest-time pattern may become inconsistent with the registered driver's normal behavior. PU behavior captures the driver's passenger-finding strategy. Different drivers may prefer different pick-up locations, operating hours, and demand-search patterns. These two behavior types provide complementary evidence: one describes rest routine, and the other describes service routine.

The IDS study converts individual driver behaviors into common and comparable features. Fisher Vector representation is used to encode STL behavior, while topic modeling is used to represent PU behavior. Self-similarity analysis compares behavioral consistency, and multi-scale pooling converts variable-length behavior evidence into fixed-length features. Multiple Component--Multiple Instance Learning (MC-MIL) is then used to handle incomplete and misaligned behavior evidence \parencite{pang2024finding}. The methodological value lies in converting personal, irregular, and noisy driver behavior into evidence that can be compared across taxis.

For transportation governance, the practical value is targeted inspection. Instead of checking taxis randomly, regulators can focus on vehicles whose behavior patterns differ from the expected routine. This can improve inspection efficiency and reduce unnecessary disturbance to compliant drivers. At the same time, the output must remain explainable: inspectors should be able to see which behavior evidence contributed to the flag, such as changed rest location, unusual working period, or altered pick-up pattern. A suspicious routine change is not proof of illegal substitution; it is a reason for further review.

\subsection{Low-Frequency Coach GPS for Unauthorized Stop Discovery}

Unauthorized coach stopping is another important regulatory scenario. In long-distance coach services, illegal passenger pick-up or drop-off may occur when a coach stops at a non-designated location. Such behavior can bypass real-name ticketing and luggage security checks, create overcrowding risks, weaken passenger rights, and increase roadside safety hazards \parencite{deng2026unsupervised}. The regulatory question is therefore not only whether a coach stopped, but whether the stop occurred at a location and pattern that deserves inspection.

The main technical challenge is low-frequency GPS. In many coach systems, GPS data may be recorded every 30 seconds or at similarly sparse intervals. A coach may slow down, stop briefly, and move again between two GPS records. Directly using zero instantaneous speed may miss many short stops. Therefore, the first task is to infer possible stop spots from sparse GPS observations.

The representative coach ASD study addresses this problem by approximating speed change within a short interval. Based on consecutive GPS points, speed, distance, and time gap, the method estimates whether a stop could have occurred between two recorded points. The inferred stop durations are then assigned to road segments, forming a stop-duration matrix:
\begin{equation}
S = [s_1; s_2; \ldots; s_M] \in \mathbb{R}^{M \times N},
\label{eq:stop_duration_matrix}
\end{equation}
where each row represents one coach journey or coach-day instance and each column represents a road segment. This matrix converts sparse trajectory evidence into a route-level behavior representation \parencite{deng2026unsupervised}.

After possible stop durations are organized, the method separates normal and abnormal stopping behavior. The central assumption is that regular stopping behavior has a shared low-rank structure, while abnormal stop evidence is sparse. A compact expression of this idea is
\begin{equation}
S = L + E,
\label{eq:low_rank_sparse_stop}
\end{equation}
where $L$ represents the low-rank regular stopping structure and $E$ represents sparse abnormal stop evidence \parencite{deng2026unsupervised}. The original model further uses low-rank, sparsity, and structured-sparsity constraints to distinguish abnormal stop spots from regular stopping patterns.

This formulation is useful for regulatory management because it produces candidate road segments rather than only a binary label for a coach. The model can highlight locations where unauthorized stops are likely to occur, allowing inspectors to allocate field resources more precisely. The study also shows that candidate abnormal spots can be interpreted through road context, such as proximity to toll stations, bus stops, railway stations, subway stations, or roadside spaces that make informal passenger pick-up convenient \parencite{deng2026unsupervised}. This improves practical interpretability because managers can connect model outputs to physical road conditions.

The governance boundary remains important. A highlighted road segment does not automatically prove that illegal passenger pick-up occurred. It indicates that stop behavior in that segment is unusual relative to normal route patterns and should be checked with supporting evidence, such as ticketing records, station logs, inspection reports, complaints, or field observations. This distinction protects both regulatory effectiveness and procedural fairness.

\subsection{Few-Shot Sparse-GPS Learning for Scalable Abnormal Stop Monitoring}

The unsupervised coach ASD model addresses the absence of labels, but many real deployments also face a different problem: a small number of confirmed abnormal cases may exist, while most data remain unlabeled. This creates a few-shot semi-supervised setting. Sparse GPS trajectories further make the problem difficult because short stop behavior may be hidden by long sampling intervals, signal noise, and incomplete motion information.

The few-shot sparse-GPS ASD study addresses these two challenges through a segment-level graph pipeline. First, Sparsity-Aware Segmentation (SAS) adaptively partitions GPS trajectories according to local spatial-temporal density. Instead of using only fixed thresholds, SAS introduces a segment break when the distance or time gap between consecutive GPS points exceeds adaptive thresholds. This preserves local behavior structure under sparse and irregular sampling \parencite{sabir2025few}.

Second, the method designs three interpretable indicators to describe abnormal stop behavior at the segment level. Temporal Influence Score (TIS) measures abnormal duration relative to local segment behavior. Maximum Speed Deviation (MSD) captures speed irregularity. Top-$k$ Aggregated Temporal Score (TTA@$k$) emphasizes dominant stop-duration evidence. These indicators are then refined by Locally Temporal-Indicator Guided Adjustment (LTIGA), which smooths indicator values over local similarity graphs to reduce the effect of sparsity and noise \parencite{sabir2025few}.

Third, the method constructs a spatial-temporal graph where each trajectory segment is a node. Label propagation expands weak supervision from a small set of labeled examples, a graph convolutional network learns relational patterns among segments, and self-training incorporates high-confidence pseudo-labels to improve robustness. The study reports that the GCN model trained after label propagation achieves an AUC of 0.8542 and AP of 0.8661, and self-training further improves performance to AUC 0.8819 and AP 0.8842 under the few-shot setting \parencite{sabir2025few}. These results are especially relevant because only a very small number of abnormal labels are available.

The main lesson for regulatory deployment is that behavior-aware indicators and graph relations can make sparse monitoring more scalable. Instead of requiring dense GPS or large labeled datasets, the method combines domain indicators, local smoothing, weak supervision, and relational learning. This is valuable for smaller agencies, intercity routes, and low-resource environments. At the same time, semi-supervised expansion must be controlled carefully. Pseudo-labels are useful for model training, but final regulatory action should still depend on human-confirmed evidence.

\subsection{Regulatory Evidence Chain for Responsible AI Use}

The three scenarios above share a common evidence-support logic. AI first converts operational records into behavior representations, then identifies suspicious changes or sparse deviations, and finally provides ranked candidates for human review. Table~\ref{tab:regulatory_evidence_chain} summarizes this evidence chain.

\begin{table}[!htbp]
\centering
\caption{Regulatory evidence-support chain for abnormal transportation behavior detection.}
\label{tab:regulatory_evidence_chain}
\scriptsize
\setlength{\tabcolsep}{3pt}
\renewcommand{\arraystretch}{1.05}
\begin{tabularx}{\textwidth}{L{2.6cm} Y Y Y Y}
\toprule
\textbf{Scenario} & \textbf{Behavioral evidence} & \textbf{AI modeling logic} & \textbf{Decision-support output} & \textbf{Governance boundary} \\
\midrule
Illegal driver substitution
& Sleeping time and location, pick-up behavior, GPS traces, taximeter records, occupied and vacant states
& Model routine consistency through STL behavior, PU behavior, self-similarity, multi-scale pooling, and MC-MIL
& Candidate taxis with suspicious routine changes for targeted inspection
& AI cannot confirm illegal substitution; final judgment requires legal evidence, driver verification, and human review \\

\midrule
Unauthorized coach stopping
& Low-frequency GPS, timestamps, speed, engine state, inferred stop duration, and road-segment location
& Infer possible stop spots and separate regular stopping behavior from sparse abnormal deviations using low-rank and sparse modeling
& Candidate abnormal stop segments for field verification and route supervision
& AI provides inspection clues only; legitimate stops, emergencies, and road conditions must be reviewed manually \\

\midrule
Sparse-GPS abnormal stop monitoring
& Sparse trajectory segments, temporal indicators, speed deviation, local density, graph relations, and limited labels
& Use SAS, TIS/MSD/TTA@$k$, LTIGA, label propagation, GCN, and self-training to learn from weak supervision
& Confidence-ranked abnormal segments for scalable monitoring
& Pseudo-labels support learning, not punishment; human-confirmed evidence and audit trails are required before regulatory action \\
\bottomrule
\end{tabularx}
\end{table}

This table emphasizes a central principle: the value of AI lies in improving the precision and efficiency of supervision, not replacing legal authority. Suspicious patterns should trigger review, not punishment. In practical systems, each flagged case should preserve a traceable path from raw data to feature construction, model output, and human decision. This audit trail is necessary for accountability, especially when model outputs affect drivers, operators, companies, or passengers.

Privacy boundaries are also critical. Driver rest locations, operating routines, income-related taximeter records, and coach movement traces can reveal sensitive personal and commercial information. Data access should therefore be limited to authorized management purposes. Aggregation, access control, anonymization, secure storage, and role-based review procedures should be used where appropriate. More importantly, data governance must follow laws, regulations, and institutional rules; technical protection only supports these requirements and cannot replace them.

Fairness should be evaluated explicitly. A model may generate more false positives for drivers or routes with irregular schedules, poor GPS quality, unusual service zones, or incomplete records. If such biases are not checked, AI-assisted inspection may unfairly burden certain drivers, companies, or regions. Therefore, abnormal behavior detection should be assessed not only by accuracy, AUC, or AP, but also by false-positive patterns, route coverage, driver-group impact, and the outcome of human review.

\subsection{Regulatory Deployment, Accountability, and Future Directions}

The management value of abnormal behavior detection lies in targeted inspection, reduced patrol cost, improved passenger safety, and fairer market order. For taxi supervision, behavior modeling can help inspectors focus on vehicles with suspicious routine changes instead of relying only on random checks. For coach regulation, abnormal stop detection can help identify locations where unauthorized pick-ups may occur and where field inspection may be most effective. For sparse-GPS monitoring, semi-supervised learning can support agencies that lack dense sensing infrastructure or large labeled datasets.

The sustainability value should be understood as an indirect management pathway. Targeted inspection can reduce unnecessary patrol mileage by enforcement vehicles. Better compliance supervision can also reduce disorderly operations, route deviations, and inefficient illegal service patterns. These effects depend on actual deployment policies, but the pathway is reasonable: behavior intelligence improves inspection precision, and improved inspection can reduce wasteful supervision while supporting safer and more orderly transport operation.

Several deployment challenges remain. The first is data sparsity. Low-frequency GPS may miss short stops, and trajectory noise can distort inferred behavior. The second is label scarcity and class imbalance. Illegal or abnormal behaviors are rare, so models must learn from limited positive examples without producing excessive false alarms. The third is interpretability. Inspectors need to understand why a taxi, coach, or road segment was flagged. Black-box outputs without behavioral explanation are weak for regulatory use. The fourth is cross-region data sharing. Coach routes may cross administrative boundaries, but data access and enforcement authority may be divided across agencies. Technical systems must therefore be supported by clear institutional agreements.

Future development should move toward accountable, human-in-the-loop regulatory intelligence. Models should produce interpretable evidence, confidence levels, case histories, and uncertainty indicators rather than only binary labels. Multi-source data, such as GPS, taximeter records, ticketing records, inspection logs, road-network context, weather, complaints, and station records, may improve robustness when used under lawful governance. Semi-supervised, few-shot, and self-supervised learning will remain important because confirmed violation labels are expensive and rare. Large models may help summarize inspection evidence or interpret textual complaints, but they should not independently determine regulatory conclusions.

AI for abnormal transportation behavior detection should therefore be developed as supervision support rather than automated enforcement. Its strongest value is to help transportation agencies find suspicious taxis, candidate abnormal coach stop locations, and weak abnormal patterns under sparse-data conditions more precisely and efficiently. Its boundary is equally clear: legal responsibility, enforcement decisions, and final judgments must remain with human authorities operating under established laws and regulations.

\section{AI for Passenger-Perceived Transit Risk Mining}
\label{sec:risk_mining}

Passenger-perceived transit risk mining focuses on discovering service problems, safety concerns, and passenger-experience signals from public digital feedback. Unlike GPS trajectories or operational records, social media posts do not directly describe vehicle movement. They describe what passengers notice, experience, complain about, and share publicly. These posts may reveal delays, crowding, fare-payment problems, unclear information, station discomfort, driver-service issues, safety concerns, or route-specific dissatisfaction. Such signals can complement structured transportation data because some service problems first appear as passenger perception before they become visible in formal reports.

The main task is not ordinary text classification. It is weak-signal discovery for service improvement. Individual posts are often short, noisy, informal, and incomplete, but repeated public expressions can reveal service risks that deserve management attention. AI can help transportation agencies organize these weak signals into interpretable topics, rank recurring concerns, and connect public feedback with service diagnosis and response. Recent reviews of topic modeling for short social media texts also emphasize that topic models should be evaluated and interpreted according to applied user needs, not only by automatic metrics \parencite{laureate2023systematic}. This point is especially important in transportation, where a topic becomes useful only when it can support service understanding, verification, and response.

\subsection{Passenger-Perceived Risk as a Complementary Sensing Layer}

Passenger-generated social media is valuable because it captures lived service experience. Operational systems may record whether a bus arrived late or whether a station was crowded, but they may not fully capture how passengers perceived the disruption. A delay may be moderate in operational terms but highly frustrating if passengers receive no explanation. A payment failure may affect only a small group of riders but still damage service trust if it repeats across stations or routes. Public posts therefore provide a complementary sensing layer for passenger experience.

However, social media is not a complete or neutral representation of all passengers. Posts are shaped by who uses the platform, who chooses to complain publicly, and whose posts receive attention. Older passengers, low-income riders, occasional users, or people without strong online presence may be underrepresented. Highly followed users may influence public attention more strongly than ordinary riders. Therefore, passenger-perceived risk mining should not treat social media as the full truth of public transport performance. It should treat it as a useful but biased source of passenger-experience evidence.

This creates two linked modeling needs. First, the model must handle short and noisy text. Transit-related posts may contain slang, emojis, route numbers, station names, abbreviations, repeated phrases, or unrelated content. Second, the model must distinguish repeated service concerns from background chatter and temporary emotional expressions. A useful system should discover topics that are both semantically meaningful and operationally interpretable.

The representative transit-risk mining study addresses this problem using city-specific Weibo posts from Beijing, Shanghai, and Xiamen. The dataset includes post text together with metadata such as likes, comments, followers, timestamps, and platform information \parencite{ashraf2025importance}. The reported corpora contain 812 posts from Beijing, 588 posts from Shanghai, and 673 posts from Xiamen, together with account-level and engagement information. This combination is important because passenger feedback has both linguistic content and social visibility. The content tells what the concern is; the metadata helps estimate how strongly the concern attracts public attention.

\subsection{Importance-Aware Topic Modeling for Weak-Signal Discovery}

A simple keyword system can identify obvious words such as ``delay,'' ``crowding,'' or ``payment,'' but it struggles with informal language and emerging issues. Classical topic models can summarize broad themes, but short social posts often provide too little context for stable topic discovery. They may also produce overlapping topics or ignore the difference between a routine low-impact post and a post that attracts wider public attention.

The representative study therefore builds an influence-weighted keyword co-occurrence graph. After cleaning and normalizing each post, adjacent keyword pairs are collected to represent local semantic relations in short text. Each post is assigned an influence weight based on engagement and reach, and each word can be assigned a salience score. The resulting edge weight between two keywords can be written as
\begin{equation}
\omega(i,j)=s_i s_j \sum_{p\in \mathcal{P}} w_p \delta_{ij}^{(p)},
\label{eq:influence_weighted_graph}
\end{equation}
where $s_i$ and $s_j$ are word salience scores, $w_p$ is the influence weight of post $p$, and $\delta_{ij}^{(p)}$ indicates whether the keyword pair appears in that post \parencite{ashraf2025importance}. This formulation allows repeated and socially visible word associations to contribute more strongly to the keyword graph.

The core model is Poisson Deconvolution Factorization. Given the influence-weighted keyword graph $W$, the model decomposes it as
\begin{equation}
W \approx UAU^{\top}
+
UH^{\top}
+
HU^{\top},
\label{eq:pdf_transit_risk}
\end{equation}
where $U$ is the topic--keyword basis, $A$ is a diagonal topic-importance matrix, and $H$ represents topic-localized residual interactions \parencite{ashraf2025importance}. The first term captures stable low-rank topical structure. The residual terms capture local associations that may involve event names, station names, route numbers, or incident-specific expressions that do not fit cleanly into broad topics.

This decomposition is useful for passenger-risk mining because social media contains both repeated service themes and localized event-specific signals. Repeated themes may include waiting, crowding, payment, route changes, vehicle service, or customer support. Local residual interactions may point to a specific route, station, incident, weather event, or safety concern. The model therefore helps separate general passenger-experience categories from more specific event descriptors.

The study also uses a decorrelation regularizer to reduce topic overlap and a coherence-driven sweep to select the number of topics. In the reported experiments, $K=10$ provides the clearest topic structure, with the highest NPMI and topic diversity among the tested topic numbers \parencite{ashraf2025importance}. In the baseline comparison, the proposed model achieves the highest NPMI and highest topic diversity, while some baselines obtain higher $C_v$ coherence. This should be interpreted carefully: the main value for this chapter is not a universal metric victory, but the ability to produce focused, diverse, and management-readable topics from noisy passenger feedback. The ablation study further shows that removing influence weighting, residual learning, or graph modulation weakens topic quality, supporting the importance of influence-aware weak-signal modeling.

\subsection{From Risk Topics to Service Feedback Governance}

The output of passenger-perceived risk mining is not only a list of words. It is a service-feedback representation that can help managers understand what passengers are reporting, where concerns may be concentrated, and which issues deserve verification. The representative study shows that learned topics and event keywords can reflect operationally meaningful concerns such as route identification, boarding and fare-payment behavior, waiting time, driver conduct, regional service variation, and safety-related incidents \parencite{ashraf2025importance}.

Topic importance is also useful. Because columns of the topic--keyword matrix are normalized, the learned diagonal values in $A$ indicate how strongly each topic contributes to the global co-occurrence structure. A high-importance topic may represent a recurrent passenger concern or a widely discussed service issue. However, topic importance should not be treated as automatic service severity. A topic may be important because it is visible online, but factual verification is still needed before management action.

Table~\ref{tab:risk_topic_governance} summarizes how the main modeling outputs can be interpreted for passenger-experience governance. The table is not a general method comparison; it connects model components to service-management meaning and fairness checks.

\begin{table}[!htbp]
\centering
\caption{Interpreting passenger-perceived risk topics for service-feedback governance.}
\label{tab:risk_topic_governance}
\scriptsize
\setlength{\tabcolsep}{3pt}
\renewcommand{\arraystretch}{1.05}
\begin{tabularx}{\textwidth}{L{2.5cm} Y Y Y Y}
\toprule
\textbf{Model output} & \textbf{Technical meaning} & \textbf{Passenger-experience signal} & \textbf{Service-management use} & \textbf{Fairness and verification check} \\
\midrule
Influence-weighted keyword edges
& Weighted co-occurrence relations between transit-related terms
& Publicly visible associations such as route--delay, station--crowding, or app--payment
& Detect recurring expressions and socially salient concerns
& Influence should guide attention but should not silence repeated low-visibility complaints \\

\midrule
Topic--keyword basis $U$
& Interpretable groups of words forming latent topics
& Broad service-risk categories such as waiting, crowding, payment, information, or safety
& Build dashboards, complaint categories, and service-diagnosis labels
& Topics should be checked for missing concerns from less visible passenger groups \\

\midrule
Topic importance $A$
& Relative contribution of each topic to the global graph structure
& Concerns that strongly organize public discussion
& Prioritize monitoring and assign topics to responsible service units
& Importance must be combined with safety relevance, recurrence, and operational evidence \\

\midrule
Residual interactions $H$
& Localized word--topic associations outside broad topic structure
& Event-specific signals involving station names, route numbers, incidents, or temporary disruptions
& Support early warning, incident verification, and targeted service response
& Localized signals require factual checking before public communication or action \\

\midrule
Representative posts and keywords
& Human-readable evidence behind each topic
& Concrete complaints, descriptions, and passenger narratives
& Help managers understand the practical meaning of each topic
& Personal information should be minimized, and examples should not expose individual passengers \\

\midrule
Topic change over time
& Change in topic strength across time windows
& Emerging, fading, or recurring public concerns
& Monitor response effectiveness and detect repeated service failures
& Temporal spikes should be compared with operational records and external events \\
\bottomrule
\end{tabularx}
\end{table}

This interpretation layer is essential. Without it, topic modeling may remain a technical exercise with limited management value. With it, the model can support a feedback loop: public posts are converted into risk topics, risk topics are verified against operational evidence, responsible departments respond, and later posts are monitored to evaluate whether passenger concerns decrease or change.

The feedback loop should always include verification. Social media can indicate perceived risk, but it is not complete factual evidence by itself. A detected topic should be checked against operational records, incident logs, hotline complaints, field reports, vehicle data, station observations, or staff feedback. This is especially important for safety-related claims, crowding reports, route-specific incidents, or accusations involving staff behavior.

\subsection{Service Feedback, Fairness, and Future Directions}

The management value of passenger-perceived risk mining lies in earlier awareness and more responsive service improvement. Repeated posts about waiting or unclear arrivals may indicate information-service weakness. Complaints about fare payment may point to ticketing-machine, app, or card-reader problems. Reports about crowding or boarding difficulty may suggest frequency mismatch, station-design issues, or poor queue management. Topic mining helps organize these concerns so that agencies can prioritize response instead of reading large volumes of posts manually.

Fairness is a central issue. Influence weighting is useful because high-engagement posts may spread quickly and affect public trust. However, influence is not the same as representativeness. A system that only follows highly visible users may overlook routine problems reported by ordinary passengers. Therefore, topic ranking should combine multiple signals: engagement, repeated evidence across users, service relevance, geographic coverage, safety implication, and consistency with operational records.

Privacy must also be protected. Public posts may contain personal names, location information, images, or sensitive descriptions. Data collection and analysis should follow platform rules, legal requirements, and institutional governance. User identities should be minimized or aggregated whenever individual identification is not necessary. The purpose should remain service improvement and public-risk monitoring, not surveillance of individual passengers.

The low-carbon value of this task should be stated carefully. Better passenger experience can make public transport more attractive, and more attractive public transport can support sustainable mobility. However, social-media risk mining alone does not directly reduce emissions. Its contribution is indirect: it helps agencies detect service problems, improve reliability and communication, and strengthen passenger trust. These improvements may support public-transport use when combined with broader service, pricing, planning, and policy measures.

Future development will likely move in three directions. First, passenger-perceived risk mining will become more multimodal by combining text, images, short videos, location tags, official notices, and operational records. Second, large language models may help summarize complaints, extract causal descriptions, and prepare draft service responses. Their outputs should be checked against evidence because they may overgeneralize or introduce unsupported explanations. Third, risk mining will become more integrated with agency workflows, connecting social-media monitoring with hotline systems, dispatching dashboards, incident records, and service-recovery processes.

Passenger-perceived transit risk mining therefore provides a complementary lens on transportation behavior. It captures what passengers notice, worry about, and publicly report. Its strongest value lies in weak-signal discovery, complaint prioritization, service diagnosis, and feedback-loop construction. Its boundary is equally important: social media does not represent all passengers, influence does not equal truth, and AI-generated topics must be verified before management action. Used responsibly, this direction can help transportation agencies build more responsive, inclusive, and trustworthy public-transit services.

\section{Integrated Framework for Transportation Behavior Intelligence and Management}
\label{sec:integrated_framework}

The preceding sections examined transportation behavior from four connected perspectives: operational prediction, mobility pattern discovery, abnormal behavior detection, and passenger-perceived risk mining. These directions differ in data source, modeling technique, and management scenario, but they share the same purpose: transforming heterogeneous transportation data into behavior intelligence that can support practical decisions. This section integrates these directions into a unified framework for transportation behavior intelligence and management.

\begin{figure}[!htbp]
\centering
\includegraphics[width=0.95\textwidth]{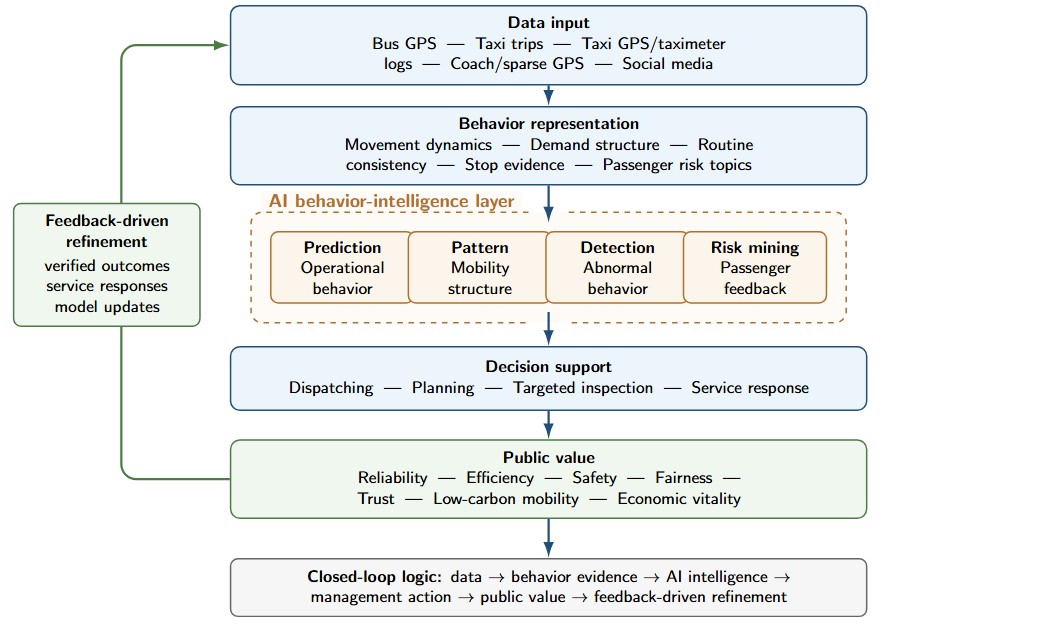}
\caption{Integrated framework for transportation behavior intelligence and management.}
\label{fig:integrated_behavior_framework}
\end{figure}

The framework is organized around a closed-loop value chain: data input, behavior representation, AI-based inference, decision support, public value, and feedback-driven refinement. This logic is shown in Figure~\ref{fig:integrated_behavior_framework}. The purpose of the framework is not to combine all models into one algorithm. Instead, it explains how different transportation data sources can be converted into behavior-aware evidence, how AI methods can infer useful intelligence from that evidence, and how the resulting outputs can support safer, more reliable, fairer, and more sustainable urban mobility management.

\subsection{From Heterogeneous Data to Behavior Intelligence}

The first layer of the framework is data input. Transportation systems generate diverse data, including bus GPS and operational records, taxi pick-up and drop-off events, taxi GPS and taximeter records, coach GPS trajectories, sparse GPS observations, and passenger-generated social media posts. These data sources differ in structure, quality, sensitivity, and management relevance. Some describe vehicle movement, some reveal travel demand, some reflect driver routines, some indicate possible compliance risks, and some capture passenger perception.

The second layer is behavior representation. This layer is necessary because raw transportation data are not directly equivalent to management knowledge. A bus GPS point becomes useful when it is represented as route progression and arrival dynamics. Taxi trip events become useful when they are represented as spatial-temporal demand patterns. Taxi operation records become useful when they reveal driver-routine consistency. Coach and sparse GPS trajectories become useful when they are transformed into stop-duration evidence and abnormal-stop indicators. Social media posts become useful when they are organized into passenger-perceived risk topics. In this sense, behavior representation is the bridge between data collection and intelligent inference.

The third layer is the AI behavior-intelligence layer. Different transportation behaviors require different AI methods. Sequential learning can support bus arrival prediction. Point-process modeling and decomposition can reveal taxi mobility structures. Behavior modeling, sparse decomposition, and graph-based learning can support abnormal behavior screening. Influence-aware topic modeling can identify passenger-perceived service risks from noisy public feedback. These methods should not be judged only by technical metrics. Their value depends on whether their outputs remain interpretable, trustworthy, and connected to management action.

\subsection{From Inference to Management Value}

The fourth layer is decision support. AI outputs become meaningful when they help transportation managers make better decisions. Bus arrival prediction can support passenger information, dispatching, interval adjustment, and route-performance diagnosis. Taxi mobility patterns can support resource allocation, hotspot monitoring, station-area planning, and multimodal coordination. Abnormal behavior detection can support targeted inspection and accountable regulatory review. Passenger-perceived risk topics can support complaint prioritization, service diagnosis, public communication, and service-recovery feedback.

The fifth layer is public value. The management value of AI-enabled transportation behavior intelligence appears in several forms. Operational prediction supports reliability and efficiency. Mobility pattern discovery supports demand matching and spatial planning. Abnormal behavior detection supports safety, accountability, and fairer regulation. Passenger-perceived risk mining supports service responsiveness and public trust. These values also connect to broader smart-city goals, including low-carbon mobility and urban economic vitality. However, these outcomes should be understood as management pathways rather than automatic effects. AI can support better decisions, but the final value depends on how agencies use model outputs, how policies are implemented, and how outcomes are evaluated.

The sixth layer is feedback-driven refinement. Transportation behavior changes over time because of route adjustment, land-use development, weather, special events, platform behavior, regulation, and passenger expectations. Therefore, AI systems should not be treated as fixed tools. Management outcomes should feed back into data collection, model calibration, and decision rules. Confirmed inspection results can improve abnormal detection. Dispatching outcomes can improve prediction evaluation. Mobility-management actions can update demand models. Passenger feedback after service changes can refine risk-topic monitoring. This feedback loop allows transportation AI to move from one-time analysis toward adaptive management intelligence.

\subsection{Deployment Pathways and Governance Principles}

The integrated framework can be deployed at different levels according to institutional readiness. A basic deployment may rely on single-source GPS or operational records with lightweight models for prediction, monitoring, or simple screening. This level is suitable for agencies with limited data infrastructure, but it still requires clear data access rules, manual verification, and transparent communication of model limitations.

A standard deployment combines multiple operational sources, such as GPS, trip records, taximeter logs, sparse trajectories, inspection records, and selected public-feedback data. At this level, agencies can build coordinated dashboards for reliability monitoring, demand analysis, abnormal stop screening, and service-risk prioritization. The main requirement is not only stronger modeling capacity, but also better data governance, traceable outputs, role-based access control, and human review procedures.

An advanced deployment integrates multimodal and cross-agency data, including operational records, road events, weather, ticketing data, complaints, social media, and verified field feedback. It may also use adaptive learning, graph-based reasoning, uncertainty estimation, and small--large model collaboration. For example, a lightweight model can perform real-time prediction, while a larger model interprets abnormal event information and provides contextual support. Such systems can support closed-loop governance, but they also require stronger accountability, privacy protection, audit trails, and regular evaluation of public value.

An additional extension of the advanced deployment level is agent-based transportation behavior intelligence. In this view, passengers, drivers, vehicles, operators, inspectors, platforms, and public agencies can be represented as interacting agents with different goals, constraints, and behavioral rules. The behavior representations discussed in this chapter can provide empirical inputs for such agents: predicted arrival dynamics can support passenger and dispatcher agents, mobility patterns can support demand and vehicle-allocation agents, abnormal-behavior evidence can support inspection agents, and passenger-risk topics can support service-response agents. Agent-based simulation can therefore help evaluate how individual decisions and management interventions may produce system-level outcomes, while keeping human authorities responsible for final policy and regulatory decisions \parencite{bastarianto2023agent}.

Across all deployment levels, trustworthy governance is the foundation. Privacy protection should begin at data collection. Fairness assessment should examine whether model benefits and errors are unevenly distributed across routes, districts, passenger groups, drivers, or operators. Interpretability should allow managers to understand why a delay warning, hotspot, suspicious stop, or risk topic was generated. Human-in-the-loop review should remain central, especially for regulatory and service-critical decisions. AI can recommend, screen, summarize, and prioritize, but final decisions must follow institutional rules, legal requirements, and professional judgment.

The integrated framework connects the chapter's technical components into one management logic. Transportation data are converted into behavior representations; AI models infer predictions, patterns, anomalies, or risks; managers use these outputs for dispatching, planning, inspection, and service improvement; and verified outcomes feed back into system refinement. This closed-loop logic reflects the central argument of the chapter: AI contributes to sustainable smart-city transportation not merely by improving isolated algorithms, but by transforming transportation behavior data into trustworthy, actionable, and value-oriented management intelligence.

\section{Challenges, Open Issues, and Future Directions}
\label{sec:challenges}

The preceding sections have shown how AI can transform transportation data into behavior intelligence for prediction, mobility pattern discovery, abnormal behavior detection, passenger-perceived risk mining, and integrated management support. However, moving from task-level AI models to reliable transportation-management systems remains challenging. The key issue is not only whether a model performs well on a dataset, but whether its output can be trusted, interpreted, verified, deployed, and connected to public value. This section discusses the main challenges and future directions that arise from the behavior-centered framework developed in this chapter.

\subsection{Behavioral Evidence Under Imperfect Data}

The first challenge is that transportation behavior is usually observed indirectly. Bus GPS records, taxi trip events, taximeter logs, coach trajectories, sparse GPS points, and social media posts are not complete descriptions of behavior. They are partial traces produced by vehicles, operators, drivers, passengers, platforms, and sensing systems. A bus delay may reflect congestion, dwell time, route design, weather, or temporary disruption. A taxi hotspot may reflect actual demand, platform behavior, or missing supply. A suspicious driver routine may indicate illegal substitution, but it may also result from shift adjustment, special demand, or incomplete records. A passenger-risk topic may reveal a real service problem, but it may also reflect temporary online attention or biased public expression.

Future transportation AI systems should therefore treat data as behavioral evidence rather than behavioral truth. Evidence quality should be assessed before model outputs are used for management. Important questions include whether the data are dense enough, whether important groups or areas are missing, whether labels are reliable, whether observations are affected by sensing noise, and whether the same pattern has alternative explanations. This is especially important in abnormal behavior detection, where confirmed violations are rare, labels are costly, and sparse GPS observations may hide short stops or distort stop duration.

Representativeness is also a major concern. Data-rich central routes may receive better prediction quality than suburban routes. Taxi records may represent only part of urban travel demand. Social media posts may overrepresent passengers who are active online and underrepresent passengers who rarely complain publicly. If these biases are ignored, AI systems may improve management only for already visible groups and places. Future work should therefore pay more attention to data coverage, weakly observed behaviors, and low-resource deployment conditions.

\subsection{Reliable, Interpretable, and Transferable Inference}

The second challenge is the gap between model performance and decision reliability. A model that works well under historical conditions may fail when traffic patterns, land use, route structures, policies, passenger demand, weather, or special events change. This issue appears across all tasks in the chapter. Bus arrival prediction must handle abnormal disruptions and route-level differences. Taxi mobility pattern discovery must distinguish stable demand structures from temporary deviations. Abnormal behavior detection must remain reliable under sparse trajectories and imbalanced labels. Passenger-perceived risk mining must separate repeated service concerns from noisy public expression.

Future models should therefore be evaluated beyond single-dataset accuracy. Transferability across routes, districts, time periods, and cities should be tested more carefully. Few-shot learning, domain adaptation, continual learning, and graph-based learning are promising, but they should not be treated as automatic solutions. A model transferred from one setting to another may carry hidden assumptions about infrastructure, regulation, passenger behavior, or data collection. Local validation and domain review remain necessary before deployment.

Interpretability is equally important. Transportation managers need to understand why a model produces a delay warning, a mobility hotspot, a suspicious stop, or a passenger-risk topic. A useful system should provide evidence that can be inspected: route progression and dwell-time information for prediction, spatial and temporal context for mobility patterns, stop-duration or routine-change evidence for abnormal behavior, and representative posts or keywords for passenger-risk topics. Without such explanation, AI outputs may be difficult to use in dispatching, planning, inspection, or service improvement.

Uncertainty should also become part of model output. A predicted arrival time should indicate confidence under abnormal conditions. A detected abnormal stop should show evidence strength and possible alternative explanations. A social-media risk topic should remain a passenger-perceived signal until it is compared with operational records or service reports. This is important because many transportation decisions affect passengers, drivers, operators, and public trust. AI should support judgment, not hide uncertainty behind a single score or label.

\subsection{Governance, Accountability, and Public Value}

The third challenge is responsible use. Transportation AI systems may influence passenger information, resource allocation, inspection priority, service response, and public communication. Therefore, AI outputs should not be treated as neutral technical facts. They are model-based interpretations of imperfect data and must be used within clear governance boundaries.

Privacy protection is fundamental. Vehicle trajectories, taximeter records, driver routines, inspection logs, and passenger-generated posts may contain sensitive information. Data use should follow legal, institutional, and public-interest requirements. Technical methods such as anonymization, aggregation, access control, differential privacy, or federated learning can support safer data use, but they do not replace governance rules. Technology should help implement privacy and compliance requirements, not define them independently. This view is consistent with trustworthy-AI guidance that treats AI risk management as a socio-technical process involving data, models, people, organizations, and deployment context \parencite{ai2023artificial}.

Fairness should also be evaluated in practical transportation terms. A prediction system should not work well only on central or high-frequency routes. A demand-discovery system should not direct all service attention to already active commercial areas while ignoring peripheral communities. An abnormal-behavior detector should not repeatedly flag operators simply because their data are noisier or their routes are less regular. A social-media mining system should not treat highly visible users as the only voice of passenger experience. Fairness therefore requires checking who benefits from AI outputs, who is exposed to model errors, and who remains underrepresented in the data.

Human accountability must remain central. In operational prediction, dispatchers should make final service decisions. In planning, AI-derived patterns should support expert and institutional discussion. In regulation, abnormal behavior detection should provide inspection clues, not legal conclusions. In passenger-risk mining, public feedback topics should guide service diagnosis but should be verified through operational evidence before major action is taken. This boundary is not a limitation of AI; it is necessary for trustworthy transportation governance.

\subsection{Future Directions for Deployable Behavior Intelligence}

Future transportation AI should move toward adaptive, multimodal, and value-oriented behavior intelligence. Adaptive systems are needed because transportation behavior changes over time. Routes are adjusted, land use develops, passenger expectations shift, platforms change service rules, and abnormal events disrupt normal patterns. Management outcomes should therefore feed back into model updating. Confirmed inspection results can improve abnormal detection. Dispatching outcomes can improve prediction evaluation. Planning actions can update mobility-demand interpretation. Passenger feedback after service changes can refine risk monitoring.

Multimodal integration is another important direction. Many transportation problems cannot be solved well from one data source alone. Bus prediction can benefit from road events, weather, signals, and passenger information. Mobility pattern discovery can be strengthened by land-use, public-transit, and event data. Abnormal behavior detection can be improved by combining GPS, ticketing records, inspection logs, complaints, and road context. Passenger-risk mining becomes more useful when social media topics are compared with operational records and service-response data. The challenge is not only technical fusion, but also lawful data sharing and institutional coordination.

A further direction is to connect behavior intelligence with agent-based transportation simulation. Agent-based models are useful because transportation outcomes often emerge from interactions among passengers, drivers, vehicles, operators, inspectors, platforms, and public agencies. In the context of this chapter, agent-based modeling should be understood as a downstream use of behavior intelligence rather than a replacement for the AI methods discussed earlier. Predicted arrival dynamics can support passenger and dispatcher agents; mobility patterns can support demand and vehicle-allocation agents; abnormal-behavior evidence can support inspection agents; and passenger-risk topics can support service-response agents. This perspective can help evaluate counterfactual management questions, such as how passengers may respond to service changes, how taxis may redistribute after demand shifts, how inspection strategies may affect abnormal operations, or how communication strategies may influence public trust \parencite{bastarianto2023agent}. However, agent-based outputs should be treated as simulation evidence, not deterministic forecasts, because agent rules, assumptions, and behavioral parameters require local calibration, validation, and auditability.

Large models and foundation models may also support future transportation behavior intelligence, but their role should be carefully constrained. They are not suitable replacements for lightweight prediction or detection models that require low latency and numerical stability. Their value is stronger in semantic and agentic decision-support tasks, such as interpreting road-event reports, summarizing complaints, extracting contextual explanations, generating scenario descriptions, and helping managers understand abnormal situations. Recent work on large language models for transportation frames these models as information processors, knowledge encoders, component generators, and decision facilitators \parencite{nie2025exploring}. In this chapter, such models should remain bounded by verified data, traceable evidence, audit trails, and human review. A practical direction is small--large model collaboration: lightweight models handle routine real-time inference, while larger models provide contextual interpretation when abnormal or information-rich scenarios occur.

Evaluation should also become more value-oriented. Prediction error, classification accuracy, AUC, clustering quality, and topic coherence remain important, but they are not enough. AI systems should also be evaluated by management usefulness, interpretability, fairness, robustness, deployment cost, service reliability, regulatory accountability, and support for low-carbon and inclusive mobility. This shift is necessary because the purpose of transportation AI is not only to improve isolated algorithms, but to support better decisions in real smart-city systems.

The future of AI-enabled transportation behavior management depends on the joint development of reliable behavioral evidence, robust inference, accountable governance, and practical deployment. AI will be most valuable when it helps managers understand transportation behavior more clearly, act more precisely, and evaluate outcomes more responsibly. Under these conditions, behavior intelligence can support safer, more reliable, more inclusive, and more sustainable transportation systems.

\section{Conclusion}
\label{sec:conclusion}

This chapter examined how artificial intelligence can support the understanding and management of transportation behavior in sustainable smart cities. The central argument is that transportation data become useful for management only when they are interpreted as behavioral evidence. Bus GPS records, taxi trip events, taximeter logs, coach and sparse GPS trajectories, and passenger-generated social media posts do not automatically provide management intelligence. They must first be transformed into meaningful representations of movement dynamics, demand structures, driver routines, stop evidence, and passenger-perceived service risks.

The chapter discussed four connected directions of AI-enabled transportation behavior intelligence. Bus arrival prediction showed how sequential learning and heterogeneous route representation can support service reliability, passenger information, dispatching, and planning feedback. Taxi mobility pattern discovery showed how point-process modeling, regularity extraction, and disparity factorization can reveal interpretable urban demand structures. Abnormal transportation behavior detection showed how driver-routine modeling, low-rank sparse stop discovery, and graph-based few-shot learning can support targeted inspection under sensitive and sparse-data conditions. Passenger-perceived transit risk mining showed how influence-aware topic modeling can organize noisy public feedback into service-risk topics and management-relevant concerns.

These directions are not isolated technical tasks. Together, they show how AI can form a bridge between transportation data and management action. Prediction helps managers anticipate operational change. Pattern discovery helps them understand spatial-temporal demand. Abnormal behavior detection helps them prioritize inspection and regulatory review. Passenger-risk mining helps them respond to public concerns and improve service feedback. In each case, the value of AI depends not only on model performance, but also on whether the output is interpretable, verifiable, and useful for real transportation decisions.

The integrated framework developed in this chapter connects these directions through a closed-loop logic: data input, behavior representation, AI behavior-intelligence inference, decision support, public value, and feedback-driven refinement. This logic is important because transportation behavior changes over time. Routes are adjusted, land use develops, passenger demand shifts, regulations evolve, and abnormal events disturb normal patterns. Therefore, AI systems should not be treated as fixed analytical tools. Their outputs and management outcomes should feed back into data collection, model calibration, service adjustment, inspection planning, and public-response evaluation.

The chapter also emphasized that transportation AI must be trustworthy and deployable. Many applications involve sensitive movement data, driver routines, passenger feedback, rare abnormal events, uneven data coverage, and decisions that may affect passengers, drivers, operators, and public agencies. For this reason, AI should remain a decision-support tool rather than an autonomous authority. Privacy protection, fairness assessment, interpretability, auditability, uncertainty awareness, and human-in-the-loop review are necessary conditions for responsible use, especially in regulatory and service-critical contexts.

Future AI-enabled transportation behavior management should move toward adaptive, multimodal, and value-oriented systems. Heterogeneous data sources should be integrated under clear governance rules. These outputs can also feed agent-based simulation and multi-agent decision-support workflows, helping agencies examine how passengers, drivers, vehicles, inspectors, platforms, and service managers may respond to policy or operational changes. Lightweight models can support real-time prediction, monitoring, and screening, while larger models may help interpret external events, summarize public feedback, and provide contextual explanations. The most promising direction is not to replace transportation managers, but to provide clearer evidence, better explanations, and more reliable support for decision-making.

Overall, AI can contribute to safer, more reliable, more inclusive, and more sustainable transportation systems when it is grounded in behavioral understanding and responsible governance. By transforming heterogeneous transportation data into trustworthy and actionable behavior intelligence, smart cities can improve operations, support planning, strengthen regulation, respond to passenger concerns, and move toward more adaptive and sustainable mobility management.

\printbibliography

@article{pang2018learning,
  title={Learning to predict bus arrival time from heterogeneous measurements via recurrent neural network},
  author={Pang, Junbiao and Huang, Jing and Du, Yong and Yu, Haitao and Huang, Qingming and Yin, Baocai},
  journal={IEEE Transactions on Intelligent Transportation Systems},
  volume={20},
  number={9},
  pages={3283--3293},
  year={2018},
  publisher={IEEE}
}

@article{pang2017discovering,
  title={Discovering fine-grained spatial pattern from taxi trips: Where point process meets matrix decomposition and factorization},
  author={Pang, Junbiao and Huang, Jing and Yang, Xue and Wang, Zuyun and Yu, Haitao and Huang, Qingming and Yin, Baocai},
  journal={IEEE Transactions on Intelligent Transportation Systems},
  volume={19},
  number={10},
  pages={3208--3219},
  year={2017},
  publisher={IEEE}
}

@article{pang2024finding,
  title={Finding a taxi with illegal driver substitution activity via behavior modelings},
  author={Pang, Junbiao and Sabir, Muhammad Ayub and Wang, Zuyun and Hu, Anjing and Yang, Xue and Yu, Haitao and Huang, Qingming},
  journal={IEEE Transactions on Intelligent Transportation Systems},
  volume={25},
  number={12},
  pages={20309--20319},
  year={2024},
  publisher={IEEE}
}

@article{deng2026unsupervised,
  title={Unsupervised Abnormal Stop Detection for Long-Distance Coaches With Low-Frequency GPS},
  author={Deng, Jiaxin and Xu, Jiayu and Pang, Junbiao and Sabir, Muhammad Ayub and Yu, Haitao},
  journal={IEEE Transactions on Intelligent Transportation Systems},
  year={2026},
  publisher={IEEE}
}

@article{sabir2025few,
  title={Few Shot Semi-Supervised Learning for Abnormal Stop Detection from Sparse GPS Trajectories},
  author={Sabir, Muhammad Ayub and Pang, Junbiao and Wu, Jiaqi and Ashraf, Fatima},
  journal={arXiv preprint arXiv:2510.12686},
  year={2025}
}

@article{ashraf2025importance,
  title={Importance-aware Topic Modeling for Discovering Public Transit Risk from Noisy Social Media},
  author={Ashraf, Fatima and Sabir, Muhammad Ayub and Deng, Jiaxin and Pang, Junbiao and Yu, Haitao},
  journal={arXiv preprint arXiv:2512.06293},
  year={2025}
}

@article{elassy2024intelligent,
  title={Intelligent transportation systems for sustainable smart cities},
  author={Elassy, Mohamed and Al-Hattab, Mohammed and Takruri, Maen and Badawi, Sufian},
  journal={Transportation Engineering},
  volume={16},
  pages={100252},
  year={2024},
  publisher={Elsevier}
}

@article{ai2023artificial,
  title={Artificial intelligence risk management framework (AI RMF 1.0)},
  author={AI, NIST},
  journal={URL: https://nvlpubs. nist. gov/nistpubs/ai/nist. ai},
  pages={100--1},
  year={2023}
}

@article{zhang2023federated,
  title={Federated learning in intelligent transportation systems: Recent applications and open problems},
  author={Zhang, Shiying and Li, Jun and Shi, Long and Ding, Ming and Nguyen, Dinh C and Tan, Wuzheng and Weng, Jian and Han, Zhu},
  journal={IEEE Transactions on Intelligent Transportation Systems},
  volume={25},
  number={5},
  pages={3259--3285},
  year={2023},
  publisher={IEEE}
}

@article{jin2023spatio,
  title={Spatio-temporal graph neural networks for predictive learning in urban computing: A survey},
  author={Jin, Guangyin and Liang, Yuxuan and Fang, Yuchen and Shao, Zezhi and Huang, Jincai and Zhang, Junbo and Zheng, Yu},
  journal={IEEE transactions on knowledge and data engineering},
  volume={36},
  number={10},
  pages={5388--5408},
  year={2023},
  publisher={IEEE}
}

@article{guo2024towards,
  title={Towards explainable traffic flow prediction with large language models},
  author={Guo, Xusen and Zhang, Qiming and Jiang, Junyue and Peng, Mingxing and Zhu, Meixin and Yang, Hao Frank},
  journal={Communications in Transportation Research},
  volume={4},
  pages={100150},
  year={2024},
  publisher={Elsevier}
}

@article{chen2024trajectory,
  title={Trajectory Data Management and Mining: A Survey from Deep Learning to the LLM Era},
  author={Chen, Wei and Zhu, Yuanshao and Chang, Yanchuan and Luo, Kang and Wen, Haomin and Li, Lei and Yu, Yanwei and Wen, Qingsong and Chen, Chao and Zheng, Kai and others},
  journal={arXiv preprint arXiv:2403.14151},
  year={2024}
}

@article{laureate2023systematic,
  title={A systematic review of the use of topic models for short text social media analysis},
  author={Laureate, Caitlin Doogan Poet and Buntine, Wray and Linger, Henry},
  journal={Artificial intelligence review},
  pages={1},
  year={2023}
}

@article{bastarianto2023agent,
  title={Agent-based models in urban transportation: review, challenges, and opportunities},
  author={Bastarianto, Faza Fawzan and Hancock, Thomas O and Choudhury, Charisma Farheen and Manley, Ed},
  journal={European Transport Research Review},
  volume={15},
  number={1},
  pages={19},
  year={2023},
  publisher={Springer}
}

@article{nie2025exploring,
  title={Exploring the roles of large language models in reshaping transportation systems: A survey, framework, and roadmap},
  author={Nie, Tong and Sun, Jian and Ma, Wei},
  journal={Artificial Intelligence for Transportation},
  volume={1},
  pages={100003},
  year={2025},
  publisher={Elsevier}
}

\end{document}